\documentclass{article}

\usepackage[misc]{ifsym}
\newcommand{\envelopefootnote}[1]{%
  \textsuperscript{\Letter}
  \begingroup
    \renewcommand{\thefootnote}{\Letter}
    \footnotetext{\scriptsize #1}
  \endgroup
}

\usepackage{colm2024_conference}

\setcitestyle{numbers,square,sort&compress}

\usepackage[table,dvipsnames]{xcolor}
\definecolor{lightgray}{gray}{0.9}
\definecolor{quotecolor}{RGB}{70,70,70}
\definecolor{lightpurple}{RGB}{230,230,250}
\definecolor{grpTextOnly}{HTML}{ECFEFF}
\definecolor{grpDetector}{HTML}{EFF6FF}
\definecolor{grpParallel}{HTML}{F0FDF4}
\definecolor{grpSerial}{HTML}{FFFBEB}
\definecolor{grpComplete}{HTML}{FAF5FF}
\definecolor{grpTiered}{HTML}{FDF2F8}

\usepackage{graphicx}
\usepackage{booktabs}
\usepackage{array}
\usepackage{tabularx}
\usepackage{colortbl}
\usepackage{multirow}
\usepackage{longtable}
\usepackage{float}
\usepackage{makecell}
\usepackage{adjustbox}
\usepackage{wrapfig}
\usepackage{caption}
\usepackage{subcaption}
\usepackage{algorithm}
\usepackage{algpseudocode}

\usepackage{amsmath, amssymb}
\usepackage{amsthm}

\usepackage{amsmath,amsfonts,bm}

\def\eqref#1{equation~\ref{#1}}

\def\1{\bm{1}}

\def\vb{{\bm{b}}}
\def\vc{{\bm{c}}}

\def\vh{{\bm{h}}}

\def\vm{{\bm{m}}}

\def\vr{{\bm{r}}}

\def\vu{{\bm{u}}}
\def\vv{{\bm{v}}}
\def\vw{{\bm{w}}}
\def\vx{{\bm{x}}}

\def\vz{{\bm{z}}}

\def\mD{{\bm{D}}}

\def\mF{{\bm{F}}}

\def\mH{{\bm{H}}}

\def\mS{{\bm{S}}}

\def\mU{{\bm{U}}}

\def\mW{{\bm{W}}}

\DeclareMathAlphabet{\mathsfit}{\encodingdefault}{\sfdefault}{m}{sl}
\SetMathAlphabet{\mathsfit}{bold}{\encodingdefault}{\sfdefault}{bx}{n}

\def\sR{{\mathbb{R}}}

\usepackage{url}
\usepackage{xurl}
\usepackage{nicefrac}
\usepackage[normalem]{ulem} 

\usepackage{CJKutf8}

\usepackage{tikz}
\usetikzlibrary{positioning, arrows.meta}
\usepackage{pgfplots}
\pgfplotsset{compat=1.18}

\usepackage{xspace}

\usepackage[raster,skins]{tcolorbox}
\tcbuselibrary{breakable}
\usepackage{transparent}

\usepackage{calligra}

\newcommand{\system}{\texttt{PGuard}\xspace}

\title{\system: A Native, Streaming, and Plug-in Guardrail for Multimodal Large Language Models}

\author{
 \vspace{15pt}
 \vspace{20pt}
 Alibaba AAIG 
 \vspace{20pt}
 }
\date{September 2026}

\begin{document}
\begin{CJK}{UTF8}{gbsn} 

\maketitle

\vspace{5pt}
\begin{abstract}
As streaming generation becomes the dominant serving paradigm for both text-only and multimodal large language models, ensuring safety throughout the generation process has become a critical challenge. 
Conventional external guardrails address this challenge at the cost of additional inference overhead and a potential gap between their safety judgments and the backbone model’s internal representations. 
Meanwhile, frontier foundation models possess increasingly sophisticated semantic understanding and rich internal representations, offering an opportunity to build safety capabilities directly into the backbone.
Motivated by this, we introduce \system, a native, streaming, and plug-in guardrail that transforms safety monitoring from an external serving dependency into an integrated capability of the backbone model.
\system directly leverages the backbone’s intermediate representations for token-level safety detection, without altering its original inference pathway or generation behavior. 
Specifically, dynamic layer gating selectively aggregates residual-stream representations across layers to extract safety-relevant signals during both prefill and autoregressive decoding.
To balance detection accuracy and computational efficiency, \system supports multiple representation-modeling architectures, including a lightweight stateless variant and sequence-aware variants built on multi-scale causal EMA-attention and GRU-based recurrent modules.
Its design also accounts for the execution characteristics of LLM serving frameworks, such as SGLang, optimizing computation and data access to minimize additional runtime and synchronization overhead. 
As a result, \system incurs only $0.82$--$2.26\%$ end-to-end overhead on the hundred-billion-parameter backbone Qwen3.5-122B-A10B served by SGLang.
Extensive evaluations across more than 30 benchmarks spanning text and multimodal safety, jailbreaks, cyber and CBRN risks, multilingual scenarios, and real-world traffic demonstrate state-of-the-art performance among native guardrails. 
On text queries with Qwen3.5-9B, \system achieves an average $F_1$ score of 0.836, approaching dedicated external guardrails, while supporting deployment-time configuration of category-specific thresholds and enforcement actions.

\end{abstract}

\section{Introduction}\label{sec:intro}
{
Large language models (LLMs) and multimodal large language models (MLLMs) are increasingly deployed in interactive applications, including agentic generation, voice chat, and incremental content rendering. 
In these settings, model responses are streamed to users as tokens are generated, rather than being presented only after generation is complete, shifting safety moderation from a post-generation task to an online streaming requirement.
Specifically, a moderation decision made only after generation completes cannot prevent harmful content that has already been exposed to users. 
Effective moderation must therefore keep pace with token generation while introducing minimal computational overhead to remain practical at serving scale.}

{
Existing guardrails address this requirement through two broad approaches. \emph{External} guardrails, such as Qwen3Guard~\citep{qwen3guard}, employ separate models to classify user inputs and model responses. 
Although they can achieve strong detection accuracy, they introduce additional inference costs. 
More importantly, post-hoc moderation can only identify unsafe content after it has been generated, whereas streaming moderation requires frequent inference by a separate model to detect risks during generation.
Furthermore, operating outside the foundation model limits their ability to directly leverage its internal representations. 
\emph{Native} guardrails instead exploit signals from the protected model itself. HiddenDetect~\citep{hiddendetect} detects input-side risks before generation, while NExT-Guard~\citep{nextguard} monitors internal features at the token level. 
Despite this progress, existing native approaches remain constrained by the effectiveness of their safety representations and the runtime cost of extracting and processing them. Our evaluations further show that existing native approaches lag behind leading external guardrails in aggregate detection performance while still incurring appreciable serving overhead.
}

{
These limitations motivate a more effective way to leverage the backbone’s internal representations for native safety modeling.
The intermediate representations of a frontier foundation model encode rich semantic and contextual information about the content being generated, providing a natural basis for identifying safety risks without duplicating much of the computation performed by a separate guardrail model. 
However, effectively translating these representations into safety signals raises two challenges. First, safety-relevant features are not uniformly distributed across layers or token positions, making adaptive layer selection important for effective representation extraction. 
Second, risks may emerge from relationships among tokens and evolve as a response unfolds, which cannot always be captured by classifying each position independently. 
Addressing these challenges requires both adaptive representation extraction and effective modeling of dependencies across token positions.
}

To address these challenges, we present \system, a native, streaming, and plug-in guardrail built on a frozen foundation model backbone. 
\system introduces dynamic layer gating to adaptively select and aggregate residual-stream representations at each token position, enabling the guardrail to capture safety-relevant features across different layers. 
Based on these representations, \system supports two complementary modeling paradigms. The \emph{stateless} variant classifies each token position independently, while the \emph{sequence} variants explicitly model dependencies across positions to capture how risks emerge and evolve during generation. 
We instantiate sequence modeling with multi-scale causal EMA-attention and GRU-based recurrent modules, offering different trade-offs between detection quality and computational cost.
To enable localized safety supervision without requiring exhaustive token-level annotations, \system adopts segment-level supervision, computing loss only at annotated positions and for tasks with available labels. 
Our experiments show that sequence modeling improves response-level detection over the stateless variant across both text and multimodal tasks.

Beyond detection efficacy, deploying native guardrails efficiently requires coordinating safety computation with the backbone’s generation process. 
Mainstream language model inference and serving engines (e.g., vLLM~\citep{vllm} and SGLang~\citep{zheng2024sglang}) coordinate batching, parallel execution, and GPU synchronization to sustain high throughput.
Accessing intermediate representations and reporting risk scores introduces additional dependencies into this execution process. 
In particular, poorly placed synchronization can stall active batches, making the resulting overhead substantially greater than the guard computation itself. \system therefore incorporates serving-engine constraints into its execution design. We integrate \system with SGLang, aligning risk computation and reporting with the engine’s scheduling mechanisms. The integration supports both stateless and sequence variants without modifying the SGLang codebase.
Beyond SGLang, we are currently working on a vLLM integration to extend the same serving-aware execution design to another widely used inference and serving engine.

We evaluate \system on more than 30 benchmarks covering text and multimodal safety, jailbreaks, cyber and CBRN risks, multilingual scenarios, and real-world traffic. 
On text-query benchmarks with {Qwen3.5-9B}, \system achieves an average $F_1$ score of 0.836, approaching that of Qwen3Guard.
On multimodal query benchmarks with Qwen3.5-9B, it achieves an average $F_1$ score of {0.741}, compared with {0.638} for HiddenDetect and {0.474} for Llama Guard 4.
On the 100B-scale Qwen3.5-122B-A10B backbone, end-to-end latency overhead remains below $1.3\%$ for stateless and $2.4\%$ for sequence variants.
These results demonstrate that native guardrails can achieve competitive detection quality while imposing limited overhead on large-scale model serving.

Our main contributions are:
\begin{itemize}
\item We propose \system, a native streaming guardrail that leverages dynamic layer gating to extract token-wise safety representations from a frozen backbone. We further introduce complementary stateless and sequence-based modeling architectures to support both independent token classification and sequential risk modeling.
\item We develop a serving-aware execution strategy that coordinates safety computation with the scheduling and execution mechanisms of model inference engines. Our SGLang integration supports both modeling variants while minimizing redundant computation and synchronization overhead, without modifying the inference engine’s codebase.
\item We conduct extensive evaluations across more than 30 text and multimodal safety benchmarks and real-world traffic. Our analysis demonstrates the benefits of sequence modeling, characterizes the trade-offs between detection quality and serving overhead, and examines the complementary roles of native screening and external adjudication.
\end{itemize}

\section{Methodology} \label{sec:method}

\begin{figure}[t]
\centering
\includegraphics[width=1.0\textwidth]{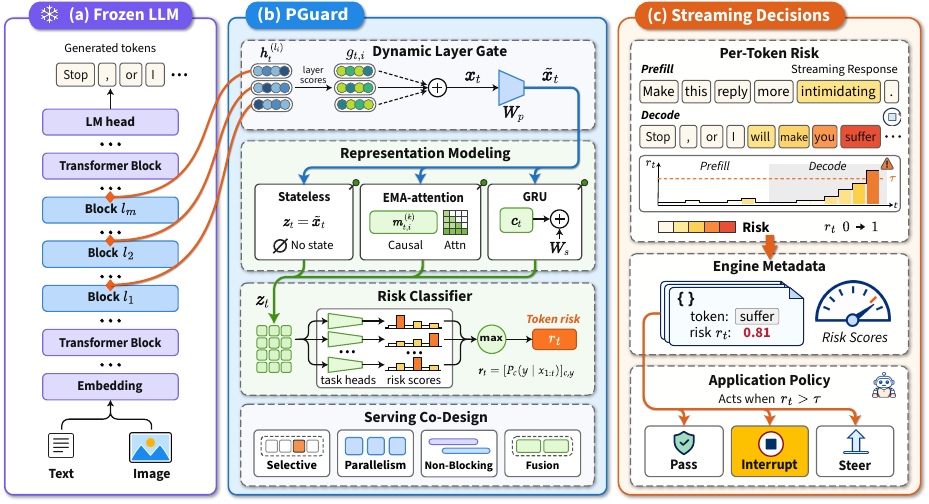}
\caption[Overview of \system.]{Overview of \system. (a) \system extracts intermediate representations from selected layers of a frozen backbone during prefill and decode. (b) Dynamic layer gating combines layer features for stateless, EMA-attention, or GRU modeling, and task-specific classifiers produce risk scores. Serving co-design integrates selective computation, parallelism, non-blocking execution, and fusion (Section~\ref{sec:impl}). (c) Per-token risk scores are returned through engine metadata, and the application applies its policy when a score crosses the threshold.}
\label{fig:architecture}
\end{figure}

\subsection{Problem Formulation and Design Goals} \label{subsec:meth_overview}

We consider a frozen backbone model, either an LLM or an MLLM, serving multi-turn conversations with text-only or multimodal inputs. A native guardrail leverages the backbone’s intermediate representations to detect risks in user inputs during prefill and in generated responses during decode. Formally, let the backbone process a token sequence $x_{1:T}$ and produce hidden states $\mH^{(l)} \in \sR^{T \times d}$ at layer $l$, where the $t$-th row $\vh^{(l)}_t \in \sR^d$ represents the hidden state at sequence position $t$. The guardrail is defined as

\[
g: \bigl(\mH^{(l_1)}, \ldots, \mH^{(l_m)}\bigr) \mapsto \bigl(\vr_1, \ldots, \vr_T\bigr), \qquad \vr_t \in [0, 1]^{|\mathcal{Y}|},
\]

which maps intermediate representations from selected backbone layers ${l_1, \ldots, l_m}$ to per-position risk scores. 
Here, $\mathcal{Y}$ denotes the shared set of risk categories across safety tasks, and $\vr_t$ contains the predicted probability of each category at position $t$. 
A position is flagged if the predicted probability of any risk category exceeds its corresponding threshold.

A production-ready native guardrail must satisfy four design goals, which guide the design of \system:

\begin{itemize}
    \item \textbf{Competitive detection performance.} The guardrail should perform competitively with dedicated external detectors.
    \item \textbf{Native integration with generation.} Risk scoring should leverage the backbone’s forward pass, enabling per-position detection during prefill and decode without modifying the backbone’s computation.
    \item \textbf{Low serving overhead.} The additional computation should contribute only a small fraction of the end-to-end serving cost, targeting a few percent overhead on production backbones.
    \item \textbf{Broad risk coverage.} A unified training framework should support text and multimodal inputs, jailbreaks, multilingual content, and emerging risks under a shared label taxonomy.
\end{itemize}

Figure~\ref{fig:architecture} illustrates the \system architecture. The frozen backbone supplies residual-stream features from selected layers. Dynamic layer gating fuses the captured features at each position, and the fused feature is modeled either by a stateless variant or by a sequence variant based on multi-scale causal EMA-attention or a GRU-based recurrent module. The risk classifiers then produce risk scores under the unified taxonomy during both prefill and decode.

\subsection{Multi-Layer Hidden State Extraction} \label{subsec:meth_feature}

Safety-relevant information is not distributed uniformly across layers. At each capture layer, \system reads the \emph{residual stream}, i.e., the tensor that enters that layer. The capture is read-only and adds no state to the frozen backbone. Training and serving capture the residual stream at matching layer boundaries, reconstructing it from separate hidden-state and residual buffers where necessary (Section~\ref{subsec:impl_sglang}).

Let $\mathcal{L} = \{l_1, \ldots, l_m\}$ denote the set of capture layers. For a sequence of $T$ positions, feature extraction produces a tensor $\mF \in \sR^{m \times T \times d}$, where $d$ is the backbone hidden size. 
In an MLLM, the visual tokens simply occupy positions in the same sequence as text tokens.
Thus, image and text evidence is fused by the backbone before feature extraction.

\subsection{Dynamic Layer Gating} \label{subsec:meth_gate}

The most discriminative safety signals can emerge at different depths across inputs and even across positions within the same input. 
To accommodate this variation, \system employs dynamic layer gating in two stages. 
During calibration-set pretraining, with the backbone frozen, the gate operates over a larger candidate set of layers, and the learned gating profiles determine the capture-layer set $\mathcal{L}$. 
During full training and deployment, the gate operates over this fixed set and adaptively weights the captured representations at each position. 
We formulate the gate over $\mathcal{L}$, with the same formulation applied to the candidate set during pretraining. 
The gate computes a score for each position-layer pair:

\begin{equation}
    s_{t,i} = \vw_g^{\top}\vh^{(l_i)}_t, \qquad
    g_{t,i}=\frac{\mathbb{1}[i\in\mathcal{I}_t]\exp(s_{t,i})}{\sum_{j\in\mathcal{I}_t}\exp(s_{t,j})},
\end{equation}
where $\vw_g$ is a learned projection and $\mathcal{I}_t$ contains the indices of the top-$L$ layers by score ranked by their scores at position $t$. During training, Gumbel noise is added to the scores before selection. 
In full training and deployment, we set $L=|\mathcal{L}|$, allowing all captured layers to participate while retaining position-specific weights. The fused representation is then computed as

\begin{equation}
    \vx_t = \sum_{i=1}^{m} g_{t,i}\, \vh^{(l_i)}_t,
\end{equation}
yielding a position-adaptive soft mixture of the captured layer representation.
This gating mechanism allows a single training recipe to accommodate backbones of different depths by learning layer preferences independently at each position (Section~\ref{subsec:exp_analysis} analyzes the learned gating profiles).

\subsection{Representation Modeling} \label{subsec:meth_head}

The backbone provides contextualized representations at each position, which \system can use either directly or through an additional module that aggregates historical evidence. 
These alternatives offer different trade-offs between detection accuracy and computational cost. 
Each variant produces a representation $\vz_t$ for the risk classifiers. Let $\tilde{\vx}_t=\mW_p\vx_t \in \sR^{d_p}$ denote the fused feature projected into a compact space, where $\mW_p \in \sR^{d_p \times d}$.

\paragraph{Stateless (last-token) variant.} 
The stateless variant independently classifies each position using its projected feature:

\begin{equation}
    \vz_t^{\mathrm{sl}} = \tilde{\vx}_t,
\end{equation}
Thus, all operations from the gate output to the logits are position-wise, with no recurrent or cached state maintained during training or inference. 
This point-wise computation makes the stateless variant the most computationally efficient option: prompt positions can be scored in parallel, while each decoding step incurs only a small additional classification cost.

\paragraph{Sequence variants: EMA-attention and GRU.} The sequence variants maintain a per-request state that accumulates safety evidence throughout generation.
We instantiate this paradigm with two designs that offer complementary trade-offs. 
The \emph{EMA-attention} design maintains causal exponential moving averages at multiple decay rates for each captured layer, indexed by $i\in\{1,\ldots,m\}$:

\begin{equation}
     \vm_{t,i}^{(k)} = \alpha_k\, \vh_t^{(l_i)} + (1 - \alpha_k)\, \vm_{t-1,i}^{(k)}, \qquad \alpha_k = 1 - e^{-1/\lambda_k}, \quad k = 1, \ldots, K,
\end{equation}
where the fixed decay $\alpha_k$ is determined by the effective length $\lambda_k$ of each scale. 
Shorter scales emphasize recent features, whereas longer scales retain a more slowly decaying summary of the earlier context. The summaries are maintained separately for each captured layer and then fused using the current gate weights, $\overline{\vm}_t^{(k)}=\sum_i g_{t,i}\vm_{t,i}^{(k)}$. 
We construct $\mS_t=[\overline{\vm}_t^{(1)},\ldots,\overline{\vm}_t^{(K)},\vx_t]$ by concatenating the fused summaries with the current feature. 
Attention then uses the current feature as the query and $\mS_t$ as both keys and values:

\begin{equation}
     \vz_t^{\mathrm{ema}} = \mW_p\,\mathrm{Attn}\bigl(Q=\vx_t,\ K=\mS_t,\ V=\mS_t\bigr),
\end{equation}

allowing each position to be represented using both its current features and a multi-scale summary of preceding context.

The \emph{GRU} design instead carries historical information in a gated recurrent unit with state $\vc_t\in\sR^{d_c}$, where $d_c$ is the state dimension. 
Since all query positions are available during prefill, their projected features $\tilde{\vx}_t$ are aggregated in parallel through attention pooling with a learned query vector.
The resulting query summary initializes the recurrent state $\vc_0$, and the classification path applied to $\vc_0$ produces the query-side prediction.
Starting from $\vc_0$, the recurrent state is updated at each response token as follows:


\begin{align}
    \vu_t &= \sigma(\mW_u \tilde{\vx}_t + \mU_u \vc_{t-1} + \vb_u), \qquad \vv_t = \sigma(\mW_v \tilde{\vx}_t + \mU_v \vc_{t-1} + \vb_v), \nonumber\\
    \hat{\vc}_t &= \tanh\bigl(\mW_c \tilde{\vx}_t + \mU_c (\vv_t \odot \vc_{t-1}) + \vb_c\bigr), \\
    \vc_t &= (1 - \vu_t) \odot \vc_{t-1} + \vu_t \odot \hat{\vc}_t, \qquad \vz_t^{\mathrm{gru}} = \vc_t + \mW_s \tilde{\vx}_t, \nonumber
\end{align}
where $\vv_t\in(0,1)^{d_c}$ is the reset gate, $\hat{\vc}_t\in\sR^{d_c}$ is the candidate state, and the update gate $\vu_t\in(0,1)^{d_c}$ weights the candidate while $1-\vu_t$ weights the previous state. The skip term $\mW_s \tilde{\vx}_t$ preserves the current position's own evidence in the output.

\subsection{Multi-Task Training} \label{subsec:meth_train}

We adopt a shared risk taxonomy adapted from YuFeng-XGuard~\citep{yufengxguard}, comprising 9 risk dimensions with 29 fine-grained categories spanning general content safety, jailbreaks, multilingual risks, and cyber \& CBRN domains. 
All training sources are mapped into this shared taxonomy (Appendix~\ref{app:taxonomy}), enabling a unified training process across different deployment scenarios.
Each task group is equipped with a multi-class classifier over its corresponding categories. The backbone remains frozen throughout training, and only the \system module is optimized.

Sample-level labels provide insufficient localization of where risks emerge within a response. 
Although token-level annotation could provide finer-grained supervision, it is expensive, while safety judgments often require interpreting spans of text in context. 
We therefore adopt \emph{segment-level supervision}, retaining each sample as a complete token sequence with multiple annotated segments. 
Each segment is annotated with labels for the relevant safety tasks, which supervise selected positions near its end. 
The classification loss is averaged over supervised positions within each segment and across available task annotations. 
For the GRU variant, the query-side prediction is additionally supervised using the label of the last query segment.

\subsection{Streaming Detection and Decision} \label{subsec:meth_decision}

\paragraph{Per-token risk and verdicts.} For each position and task $c$, the classifier outputs a probability distribution $P_c(\cdot \mid x_{1:t})$ over the task-specific categories $\mathcal{Y}_c \subseteq \mathcal{Y}$. 
A position is flagged for an unsafe category $y \in \mathcal{Y}_c$ when $P_c(y \mid x_{1:t}) \geq \tau_{c,y}$. 
Each trained \system model uses its own calibrated thresholds, configured separately for query and response stages and for text-only and image-text inputs (Appendix~\ref{app:thresholds}).
During prefill, the query-side verdict is derived from scores at selected positions or their aggregation, depending on the application's policy, and is available before response generation begins. 
During decode, the hidden state at position $t$ is used to assess the content up to and including token $x_t$, while the same forward pass predicts the next token $x_{t+1}$. 
Verdict consumption is configurable at deployment: applications can selectively enable task groups and determine how flagged content is handled, including reporting, interrupting, or steering generation.

\paragraph{Sampled scoring.} 
To further reduce serving overhead, \system supports sampled scoring, exploiting the observation that safety risks typically manifest across phrases or passages rather than in isolated tokens. 
Under this option, \system performs classification and reports risk scores only once every $p$ tokens, skipping the classification path at intermediate steps (Section~\ref{subsec:impl_coopt}). 
For sequence variants, however, the recurrent or attention-based state is still updated at every token, ensuring that each reported score incorporates evidence from the entire processed prefix. 
The early-interception analysis in Section~\ref{subsec:exp_streaming} uses the full per-token scoring trajectory rather than sampled scores.

\paragraph{Two-tier deployment.} 
\system also supports a two-tier cascade in which the native variant performs token-level scoring on all traffic, while only flagged content is escalated to an external reasoning guardrail for fine-grained adjudication (Section~\ref{subsec:exp_cascade}).

\section{System Implementation} \label{sec:impl}

\subsection{Inference Engine Integration} \label{subsec:impl_sglang}

\begin{figure}[t]
\centering
\includegraphics[width=0.60\textwidth]{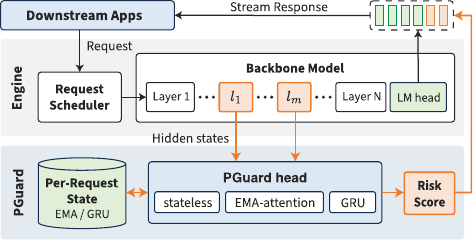}
\caption{\textbf{\system integrated into the SGLang serving engine.} \system runs as a side path of the backbone forward pass. Hidden states from the selected layers are sent to the \system head, which reads and updates a per-request state for the EMA-attention and GRU variants. The resulting risk scores are streamed to downstream applications together with the generated tokens.}
\label{fig:impl_overview}
\end{figure}

We integrate \system into SGLang~\citep{zheng2024sglang}, with runtime support for feature capture, per-request state management, and streaming risk-score delivery. The integration supports the stateless, EMA-attention, and GRU variants through a shared engine interface. Figure~\ref{fig:impl_overview} illustrates the integration.

At each selected layer, the runtime captures the same residual-stream representation used during training (Section~\ref{subsec:meth_feature}).
For backbones that maintain hidden-state and residual tensors separately, the runtime reconstructs this representation by summing the two tensors.
This operation is performed solely for feature extraction and does not alter the backbone’s forward computation. The resulting risk scores are exposed as response metadata, leaving the original token logits and sampling path unchanged.

For sequence-based variants, the runtime maintains persistent state tensors for each active request in GPU memory.
Each request is assigned a dedicated slot throughout its lifetime, enabling state retrieval independently of changes in batch composition introduced by the scheduler.
During both chunked prefill and auto-regressive decoding, the runtime retrieves the state associated with the request, updates it with newly processed features, and writes the updated state back to the corresponding slot.
For the GRU variant, prefill does not process query tokens recurrently.
Instead, it projects all query features in parallel and aggregates them into the initial state $\vc_0$ using learned-query attention pooling.
The runtime stores $\vc_0$ as the per-request state. During decode, each response token updates this state incrementally.

\paragraph{Runtime Support for \system Variants.}
All variants share the same engine interface but differ in state management.
The stateless variant requires no persistent state.
The EMA-attention variant maintains a set of multi-scale states for each request and captured layer.
During chunked prefill, each chunk consumes the final state from the preceding chunk and produces an updated state for the next chunk. During decode, these states are updated incrementally at each token.

The GRU variant maintains a single fixed-size state per request.
During prefill, learned-query attention pooling computes the initial state from the query features (Section~\ref{subsec:meth_head}), which is then stored in the request’s assigned slot.
During decode, each response token updates this state before optional scoring.
Consequently, tokens for which no risk score is reported are still incorporated into the recurrent state, preserving the complete response history.

\system also supports pipeline and tensor parallelism. Under pipeline parallelism, the Qwen3.5 integration forwards captured features in a fixed layer order and executes the guard at the final stage. Under tensor parallelism, each rank independently computes scores for a distinct subset of sampled prefill positions.
The ranks then aggregate their local maximum scores using an all-reduce operation.

\subsection{Algorithm--System Co-Optimization} \label{subsec:impl_coopt}
\label{subsec:impl_overhead}

Directly integrating a native guardrail into the backbone forward pass can increase serving latency, even when the guard head is lightweight. 
The overhead arises not only from risk-score computation but also from repeated kernel launches, sequential state updates, and host--device synchronization.

Stage-level profiling reveals two primary sources of overhead. The first is \emph{redundant work}. 
Under \system{}'s sampled reporting policy, risk scores are consumed only at selected positions, yet naively executing the guard at every token incurs unnecessary computation.
The second is \emph{inefficient execution}. Sequential state updates constrain prefill parallelism, small operators trigger repeated kernel launches, and host--device synchronization can stall the submission of subsequent backbone operations.

We address these bottlenecks through coordinated algorithmic and runtime optimizations. 
Specifically, we decouple state updates from risk scoring by computing risk scores only at positions selected for reporting, parallelize guard-state computation during prefill, move synchronization off the serving critical path, and fuse compatible operations to reduce kernel-launch overhead.

\paragraph{Selective Guard Scoring.}
Under a sampled reporting policy, applications consume a risk score every $p$ tokens. Computing scores at intermediate positions therefore incurs unnecessary overhead. 
We eliminate this redundant computation by decoupling state updates from risk scoring.
For the EMA-attention and GRU variants, every token continues to update the corresponding state to preserve the sequence context, whereas the scoring path is executed only at positions selected for reporting.

\begin{figure}[t]
\centering
\includegraphics[width=0.65\textwidth]{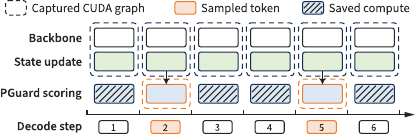}
\caption{\textbf{Selective guard scoring during decode.} The captured decode graph is split into segments. The backbone and the state update are replayed at every step. The \system scoring segment is replayed only at the sampled steps (steps 2 and 5 in this example) and is skipped at all other steps.}
\label{fig:impl_selective_scoring}
\end{figure}

SGLang captures backbone and guard operations into a single CUDA graph. 
As a result, each graph replay executes the entire scoring path, even at positions where the resulting scores are not reported. 
To avoid this, we split the decode graph into two segments. The first segment covers the backbone forward pass and the sequence-state update, and is replayed at every step. The second segment covers the guard scoring path, and is replayed only at the sampled steps (Figure~\ref{fig:impl_selective_scoring}). With a reporting interval of $p$ tokens, this removes approximately $1-1/p$ of the scoring executions.

\paragraph{Parallel Prefill Execution.}
Prefill is the primary source of \system{}'s overhead in our workloads. While the backbone processes prompt tokens in parallel, a straightforward EMA implementation updates the guard state sequentially, one token at a time.
These sequential updates limit parallelism and incur repeated kernel-launch overhead.
We exploit the linear structure of the EMA update to compute states within each chunk in parallel. For a fixed decay rate $\alpha_k$, the state at position $t$ can be expressed in terms of the state at an earlier position $s$ as follows.

\begin{equation}
    \vm_{t,i}^{(k)}
    =
    (1-\alpha_k)^{t-s}\,\vm_{s,i}^{(k)}
    +
    \sum_{j=s+1}^{t}
    \alpha_k (1-\alpha_k)^{t-j}\,\vh_j^{(l_i)} .
\end{equation}

\begin{figure}[!t]
\centering
\includegraphics[width=\textwidth]{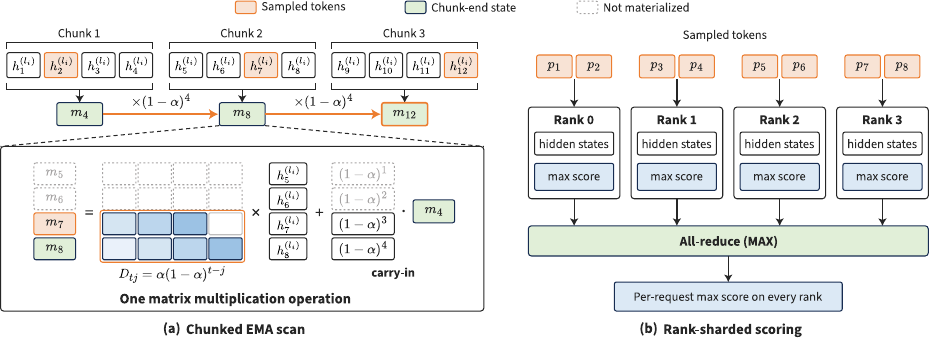}
\caption{\textbf{Parallel prefill execution.} \textbf{(a) Chunked EMA scan.} Within each chunk, the EMA states are computed by one matrix multiplication with a lower-triangular decay matrix $\mD$. The state at the end of each chunk is carried into the next chunk. States are computed only at the sampled tokens and at chunk ends. \textbf{(b) Rank-sharded scoring.} Each tensor-parallel rank scores a disjoint subset of the sampled tokens and computes a local maximum. One all-reduce with maximum reduction then gives every rank the per-request maximum risk score.}
\label{fig:impl_parallel_prefill}
\end{figure}

We partition the prompt into fixed-size chunks and represent the weighted sums using a lower-triangular decay matrix as shown in Figure~\ref{fig:impl_parallel_prefill}(a). 
Multiplying its selected rows by the token features computes the corresponding EMA summaries in parallel. 
Importantly, every prompt token contributes to these summaries, including tokens at positions where no risk score is reported. We compute the states at sampled positions and the final prompt token for scoring, and retain each chunk's final state to initialize the next chunk.

Tensor parallelism provides an additional opportunity to distribute prefill scoring as illustrated in Figure~\ref{fig:impl_parallel_prefill}(b). 
When each rank independently evaluates the guard on replicated features, the same scores are redundantly computed across ranks. 
Instead, we evenly partition the sampled positions across the $N$ ranks, with each rank computing a local maximum score for every risk category over its assigned positions. A single element-wise maximum all-reduce then aggregates these local results into the per-request score vector. 
Only the score vectors are exchanged, avoiding the transfer of hidden states across ranks. This partitioning reduces the scoring workload on each rank by approximately a factor of $N$.

\paragraph{Non-Blocking Guard Execution.}
Feature aggregation, context modeling, and score delivery introduce additional data dependencies when a native guardrail is integrated into the serving runtime. Resolving these dependencies through host--device synchronization can delay the submission of subsequent backbone operations. 
We keep state management and score aggregation on the GPU and use asynchronous transfers to reduce synchronization on the serving critical path. During prefill, fixed-interval sampling decisions are staged in pinned host memory and transferred asynchronously, with the sampled-token mask constructed on the GPU. Per-request state lookup and reset also remain on the GPU. 
Together, these changes avoid unnecessary host--device synchronization during prefill.

During decode, we overlap risk-score delivery with ongoing engine execution and align result consumption with the engine's existing synchronization schedule. Asynchronous staging in pinned host memory supports this overlap without introducing an additional synchronization barrier for score delivery.

\paragraph{Fused Guard Computation.}
The guard head consists of many small matrix operations, each requiring a separate GPU kernel launch and memory access. 
For small inputs, these overheads can dominate the actual computation and result in poor GPU utilization. 
To address this, we fuse compatible operations into larger matrix multiplications. In the multi-task classifier, all nine task heads share the same input features. We concatenate the weights of their first layers into a single matrix and arrange the weights of their output layers into a block-diagonal matrix. 
This enables all nine heads to be evaluated using only two matrix multiplications, rather than eighteen.
The fused weights are constructed once during model loading, eliminating the need for weight assembly during serving. 
This design reduces kernel-launch overhead and improves GPU utilization.

Together, these optimizations align guard execution with application reporting requirements while preserving complete sequence-state updates. 
By exploiting parallelism, reducing synchronization, and consolidating small operations, they mitigate the computational and runtime overhead of native guardrail integration along the serving critical path.

\section{Experiments} \label{sec:experiments}

\subsection{Experimental Setup} \label{subsec:exp_setup}

\subsubsection{Benchmarks} \label{subsubsec:exp_benchmark}

We  {evaluate \system across four dimensions, following the evaluation} settings of WildGuard~\cite{wildguardmix} and Qwen3Guard~\cite{qwen3guard}.

\begin{itemize}
    \item \textbf{Safety classification.} ToxicChat~\cite{toxicchat}, XSTest~\cite{xstest}, OpenAI Moderation~\cite{openaimod}, Aegis~2.0~\cite{aegis2}, WildGuardMix~\cite{wildguardmix}, S-Eval~\cite{seval}, BeaverTails~\cite{beavertails}, Qwen3Guard test set~\cite{qwen3guard}, FineHarm~\cite{scm}, and SafeRLHF~\cite{saferlhf} (text). SPA-VL~\cite{spavl}, UnsafeBench~\cite{unsafebench}, MM-SafetyBench~\cite{mmsafetybench}, VLGuard~\cite{vlguard}, HoliSafe~\cite{holisafe}, VLSBench~\cite{vlsbench}, FigStep~\cite{figstep}, and Beavertails-V~\cite{beavertails-v} (image-text).
    \item \textbf{Frontier risk classification.} StrongREJECT~\cite{strongreject}, S-Eval jailbreak subset~\cite{seval} (text), FigStep~\cite{figstep} and MM-SafetyBench~\cite{mmsafetybench} (image-text) for jailbreak detection; WMDP~\cite{wmdp}, Meta CyberSecEval~3~\cite{cyberseceval3}, and SOSBench~\cite{sosbench} for cybersecurity and CBRN risk identification.
    \item \textbf{Multilingual safety.} PolyGuardPrompts~\cite{polyguard} (17 languages), RTP-LX~\cite{rtplx} (12 languages), and Lingua-SafetyBench~\cite{lingua_safetybench} (10 languages).
\item \textbf{Over-refusal and real-world data.} XSTest~\cite{xstest} and OR-Bench~\cite{orbench}  {assess over-refusal. Benign-traffic evaluation uses Alpaca~\cite{alpaca} and WildChat~\cite{wildchat} for text, and} ShareGPT4V~\cite{sharegpt4v}, LLaVA-Instruct~\cite{llavaisa}, and MMStar~\cite{mmstar}  {for image-text inputs}.
\end{itemize}

\subsubsection{Baselines} \label{subsubsec:exp_baselines}

We compare \system  with native and external guardrails. Native guardrails derive safety signals from the protected backbone; external guardrails run a separate safety model over the query and response.

\begin{itemize}
    \item \textbf{Native guardrails}: GradSafe~\cite{gradsafe} (Llama-2-7B~\cite{llama2}), SIREN~\cite{siren} (Llama-3.1-8B~\cite{llama3}), NExT-Guard~\cite{nextguard} (Qwen3-8B), HiddenDetect~\cite{hiddendetect} (LLaVA-1.6-7B~\cite{llava16}).
    \item \textbf{External guardrails}: {SCM-7B~\cite{scm} (Qwen2.5-7B~\cite{qwen25}),} Qwen3Guard-Stream-8B~\cite{qwen3guard} (Qwen3-8B~\cite{qwen3}), Qwen3Guard-Gen-8B~\cite{qwen3guard}, WildGuard-7B~\cite{wildguardmix}, PolyGuard-Qwen-7B~\cite{polyguard},
    {YuFeng-XGuard-8B~\cite{yufengxguard},} LlamaGuard 4 (12B)~\cite{llamaguard4}, LlavaGuard-7B~\cite{llavaguard}.
\end{itemize}

\subsubsection{Implementation Details} \label{subsubsec:exp_imp}
 {Sections~\ref{subsec:exp_main}--\ref{subsec:exp_benign} evaluate \system on Qwen3-8B and Qwen3.5-9B~\cite{qwen35}. Section~\ref{subsec:exp_backbones} extends the evaluation to {Qwen3.5-35B-A3B (also referred to as Qwen3.5-Flash) and Qwen3.5-122B-A10B}.
We evaluate query and response classification against sample-level benchmark labels. For these classification evaluations, we use a scoring interval of $p=10$ tokens. Query classification uses the score} {at the query boundary}{, and response classification uses the score from sampled token scores. Decision thresholds are calibrated on a held-out validation set}.

 {We evaluate Qwen3Guard-Gen and Qwen3Guard-Stream in \texttt{strict} mode, treating the Controversial label as unsafe. Other baselines use} their officially released weights and default settings where available. HiddenDetect and NExT-Guard additionally require calibration because official settings are unavailable.

\subsection{Main Results} \label{subsec:exp_main}

\paragraph{Text Safety Evaluation}
Tables~\ref{tab:text-query} and~\ref{tab:text-response} show that \system achieves the highest average text-safety $F_1$ among the evaluated native guardrails. For queries, \system-ema (Qwen3.5-9B) and \system-gru (Qwen3.5-9B) both achieve 0.836, exceeding prior native baselines by 4.5--9.5 percentage points. This score is comparable to Qwen3Guard-Stream (0.834) and higher than Qwen3Guard-Gen (0.814). \system-ema (Qwen3-8B) achieves 0.827, falling 0.7 percentage points below Qwen3Guard-Stream but outperforming Qwen3Guard-Gen by 1.3 percentage points, both built on Qwen3-8B.
For responses, \system-gru (Qwen3.5-9B) leads the \system variants at 0.860, below Qwen3Guard-Gen-8B (0.866) and YuFeng-XGuard-8B (0.881). On Qwen3-8B, \system-ema achieves 0.858, exceeding SCM-7B by 7.6 percentage points and exceeding Qwen3Guard-Stream-8B (0.830).

Sequence variants consistently outperform the stateless variant in average $F_1$ on both query and response safety across both backbones, with gains ranging from {0.3 to 6.5} percentage points.

\begin{table}[t]
\centering
\caption{\textbf{Query} safety evaluation on \textbf{text} benchmarks. OAI-Mod denotes OpenAI Moderation; WGMix denotes WildGuardMix. The highest $F_1$ in each column is shown in \textbf{bold}.}
\label{tab:text-query}
\resizebox{0.9\textwidth}{!}{%
\begin{tabular}{@{}lccccccc@{}}
\toprule
\textbf{Model} & \textbf{ToxicChat} & \textbf{XSTest} & \textbf{OAI-Mod} & \textbf{Aegis2} & \textbf{WGMix} & \textbf{S-Eval} & \textbf{Avg} \\
\midrule
\multicolumn{8}{@{}l}{\textit{External guardrails}} \\
Qwen3Guard-Gen-8B & 0.636 & 0.908 & 0.682 & 0.865 & \textbf{0.887} & \textbf{0.906} & 0.814 \\
Qwen3Guard-Stream-8B & 0.715 & 0.921 & 0.740 & \textbf{0.869} & 0.871 & 0.890 & 0.834 \\
WildGuard-7B & 0.651 & 0.932 & 0.724 & 0.831 & 0.869 & 0.854 & 0.810 \\
PolyGuard-Qwen-7B & 0.677 & 0.923 & 0.732 & 0.866 & 0.877 & 0.898 & 0.829 \\
YuFeng-XGuard-8B & 0.705 & \textbf{0.944} & 0.720 & 0.859 & 0.866 & \textbf{0.906} & 0.833 \\
\midrule
\multicolumn{8}{@{}l}{\textit{Native guardrails}} \\
GradSafe (Llama-2-7B) & 0.733 & 0.897 & 0.698 & 0.803 & 0.761 & 0.686 & 0.763 \\
SIREN (Llama-3.1-8B) & 0.610 & 0.771 & \textbf{0.844} & 0.842 & 0.822 & 0.558 & 0.741 \\
NExT-Guard (Qwen3-8B) & \textbf{0.750} & 0.844 & 0.730 & 0.780 & 0.782 & 0.858 & 0.791 \\
\midrule
\system-lasttok (Qwen3-8B) & 0.737 & 0.886 & 0.747 & 0.837 & 0.846 & 0.889 & 0.824 \\
\system-gru (Qwen3-8B) & 0.736 & 0.869 & 0.782 & 0.838 & 0.860 & 0.894 & 0.830 \\
\system-ema (Qwen3-8B) & 0.697 & 0.928 & 0.755 & 0.853 & 0.853 & 0.876 & 0.827 \\ 
\system-lasttok (Qwen3.5-9B) & 0.692 & 0.899 & 0.767 & 0.846 & 0.846 & 0.879 & 0.822 \\
\system-gru (Qwen3.5-9B) & 0.745 & 0.887 & 0.782 & 0.848 & 0.852 & 0.903 & \textbf{0.836} \\
\system-ema (Qwen3.5-9B) & 0.716 & 0.907 & 0.781 & 0.853 & 0.858 & 0.899 & \textbf{0.836} \\ 
\bottomrule
\end{tabular}%
}
\end{table}

\begin{table}[t]
\centering
\caption{\textbf{Response} safety evaluation on \textbf{text} benchmarks. Q3G-Test denotes the Qwen3Guard test set; WGMix denotes WildGuardMix. The highest $F_1$ in each column is shown in \textbf{bold}.}
\label{tab:text-response}
\resizebox{\textwidth}{!}{%
\begin{tabular}{@{}lccccccccc@{}}
\toprule
\textbf{Model} & \textbf{BeaverTails} & \textbf{Q3G-Test} & \textbf{FineHarm} & \textbf{SafeRLHF} & \textbf{Aegis2} & \textbf{WGMix} & \textbf{XSTest} & \textbf{S-Eval} & \textbf{Avg} \\
\midrule
\multicolumn{10}{@{}l}{\textit{External guardrails}} \\
Qwen3Guard-Gen-8B & \textbf{0.865} & \textbf{0.924} & 0.972 & 0.909 & \textbf{0.863} & 0.781 & 0.827 & 0.785 & 0.866 \\
Qwen3Guard-Stream-8B & 0.851 & 0.803 & 0.960 & 0.930 & 0.827 & 0.746 & 0.890 & 0.633 & 0.830 \\
WildGuard-7B & 0.841 & 0.868 & \textbf{0.991} & 0.925 & 0.838 & 0.721 & \textbf{0.912} & 0.429 & 0.816 \\
PolyGuard-Qwen-7B & 0.802 & 0.911 & 0.983 & 0.894 & 0.824 & 0.764 & 0.349 & 0.622 & 0.769 \\
YuFeng-XGuard-8B & 0.837 & 0.907 & 0.971 & \textbf{0.950} & 0.802 & \textbf{0.782} & 0.885 & \textbf{0.910} & \textbf{0.881} \\
{SCM-7B (Qwen2.5-7B)} & {0.796} & {0.868} & {0.973} & {0.901} & {0.777} & {0.703} & {0.590} & {0.646} & {0.782} \\
\midrule
\multicolumn{10}{@{}l}{\textit{Native guardrails}} \\
SIREN (Llama-3.1-8B) & 0.842 & 0.902 & 0.869 & 0.917 & 0.822 & 0.589 & 0.591 & 0.365 & 0.737 \\
NExT-Guard (Qwen3-8B) & 0.766 & 0.849 & 0.745 & 0.800 & 0.779 & 0.404 & 0.179 & 0.546 & 0.634 \\
\midrule
\system-lasttok (Qwen3-8B) & 0.760 & 0.827 & 0.923 & 0.862 & 0.804 & 0.695 & 0.824 & 0.826 & 0.815 \\
\system-gru (Qwen3-8B) & 0.804 & 0.882 & 0.941 & 0.871 & 0.800 & 0.751 & 0.845 & 0.867 & 0.845 \\
\system-ema (Qwen3-8B) & 0.834 & 0.908 & 0.943 & 0.927 & 0.809 & 0.777 & 0.796 & 0.870 & 0.858 \\ 
\system-lasttok (Qwen3.5-9B) & 0.751 & 0.825 & 0.914 & 0.845 & 0.755 & 0.681 & 0.785 & 0.803 & 0.795 \\
\system-gru (Qwen3.5-9B) & 0.835 & 0.906 & 0.948 & 0.892 & 0.832 & 0.734 & 0.855 & 0.881 & 0.860 \\
\system-ema (Qwen3.5-9B) & 0.780 & 0.867 & 0.920 & 0.866 & 0.788 & 0.744 & 0.853 & 0.849 & 0.833 \\ 
\bottomrule
\end{tabular}%
}
\end{table}

\paragraph{Multimodal Safety Evaluation}
The same training recipe applies to multimodal backbones, which fuse visual and textual information before \system extracts their representations. In Table~\ref{tab:vlm-merged}, \system-ema (Qwen3.5-9B) achieves the highest average query $F_1$ of 0.741, compared with 0.638 for HiddenDetect and 0.474 for Llama~Guard~4. HiddenDetect remains stronger on MM-SafetyBench and FigStep. For responses, \system-gru leads the evaluated methods with an average $F_1$ of 0.798.

\begin{table}[t]
\centering
\caption{Multimodal safety evaluation. The highest $F_1$ in each column is shown in \textbf{bold}.}
\label{tab:vlm-merged}
\resizebox{\textwidth}{!}{%
\begin{tabular}{@{}lccccccccc|ccc@{}}
\toprule
 & \multicolumn{9}{c}{\textbf{Query side} $F_1$} & \multicolumn{3}{c}{\textbf{Response side} $F_1$} \\
\cmidrule(lr){2-10}\cmidrule(l){11-13}
\textbf{Model} & \textbf{SPA-VL} & \textbf{UnsafeB} & \textbf{MM-SB} & \textbf{VLGuard} & \textbf{HoliSafe} & \textbf{VLSB} & \textbf{FigStep} & \textbf{BeaverT-V} & \textbf{Avg} & \textbf{SPA-VL} & \textbf{BeaverT-V} & \textbf{Avg}\\
\midrule
\multicolumn{13}{@{}l}{\textit{External guardrails}} \\
Llama~Guard~4 (12B)            & 0.535 & 0.326 & 0.608 & 0.650 & 0.337 & 0.129 & 0.779 & 0.426 & 0.474 & 0.495 & 0.515 & 0.505 \\
LlavaGuard-7B                  & 0.256 & \textbf{0.582} & 0.269 & 0.557 & 0.424 & 0.641 & 0.224 & 0.535 & 0.436 & -- & -- & -- \\
\midrule
\multicolumn{13}{@{}l}{\textit{Native guardrails}} \\
HiddenDetect (Llava-1.6-7B)    & 0.658 & 0.234 & \textbf{0.779} & 0.798 & 0.649 & 0.215 & \textbf{0.927} & \textbf{0.845} & 0.638 & -- & -- & -- \\
\midrule
\system-lasttok (Qwen3.5-9B)    & 0.694 & 0.384 & 0.649 & 0.647 & 0.410 & 0.586 & 0.882 & 0.828 & 0.635 & 0.595 & 0.692 & 0.644 \\
\system-gru (Qwen3.5-9B)        & \textbf{0.814} & 0.448 & 0.706 & \textbf{0.837} & \textbf{0.732} & 0.678 & 0.840 & 0.822 & 0.735 & \textbf{0.800} & \textbf{0.796} & \textbf{0.798} \\
\system-ema (Qwen3.5-9B)        & 0.797 & 0.516 & 0.682 & 0.800 & 0.711 & \textbf{0.763} & 0.857 & 0.798 & \textbf{0.741} & 0.757 & 0.788 & 0.772 \\ 
\bottomrule
\end{tabular}%
}
\end{table}


\paragraph{Adversarial Jailbreak Detection.}
Table~\ref{tab:jailbreak} evaluates detection on the static jailbreak benchmarks. \system-ema (Qwen3-8B) achieves an average $F_1$ of 0.946,  ranking first among native guardrails and second overall. It trails YuFeng-XGuard-8B (0.972) and exceeds Qwen3Guard-Gen (0.928). On S-Eval-JB, its $F_1$ of 0.896 exceeds NExT-Guard on the same backbone by 12.5 percentage points, while YuFeng-XGuard-8B leads at 0.944. These results support the effectiveness of internal representations for detecting the evaluated jailbreak prompts.

\begin{table}[t]
\centering
\caption{Safety evaluation results on static jailbreak benchmarks.}
\label{tab:jailbreak}
\small
\resizebox{0.6\textwidth}{!}{%
\begin{tabular}{@{}lccc@{}}
\toprule
\textbf{Model} & \textbf{StrongREJECT} & \textbf{S-Eval-JB} & \textbf{Avg} \\
\midrule
\multicolumn{4}{@{}l}{\textit{External guardrails}} \\
Qwen3Guard-Gen-8B & 0.998 & 0.858 & 0.928 \\
Qwen3Guard-Stream-8B & 0.997 & 0.804 & 0.901 \\
WildGuard-7B & 0.995 & 0.785 & 0.890 \\
PolyGuard-Qwen-7B & 0.995 & 0.857 & 0.926 \\
YuFeng-XGuard-8B & \textbf{1.000} & \textbf{0.944} & \textbf{0.972} \\
\midrule
\multicolumn{4}{@{}l}{\textit{Native guardrails}} \\
GradSafe (Llama-2-7B) & 0.995 & 0.383 & 0.689 \\
SIREN (Llama-3.1-8B) & 0.970 & 0.396 & 0.683 \\
NExT-Guard (Qwen3-8B) & 0.990 & 0.771 & 0.881 \\
\midrule
\system-lasttok (Qwen3-8B) & 0.990 & 0.821 & 0.906 \\
\system-gru (Qwen3-8B) & 0.984 & 0.876 & 0.930 \\
\system-ema (Qwen3-8B) & 0.995 & 0.896 & 0.946 \\ 
\system-lasttok (Qwen3.5-9B) & 0.985 & 0.849 & 0.917 \\
\system-gru (Qwen3.5-9B) & 0.989 & 0.884 & 0.937 \\
\system-ema (Qwen3.5-9B) & 0.998 & 0.879 & 0.939 \\ 
\bottomrule
\end{tabular}
}
\end{table}


\paragraph{Cyber \& CBRN Risk Identification.}
In Table~\ref{tab:cyber}, \system-lasttok (Qwen3.5-9B) achieves the highest average $F_1$ of 0.833, followed by the GRU (0.789) and EMA-attention (0.773) variants on that backbone. On Qwen3-8B, \system-ema reaches 0.747, exceeding the evaluated native baselines and both Qwen3Guard variants. Performance varies across risk categories: SIREN leads on WMDP biology and chemistry, whereas \system leads on several CyberSecEval3 and SOSBench subsets.

\begin{table}[t]
\centering
\caption{\textbf{Query} safety evaluation results on Cyber \& CBRN benchmarks.}
\label{tab:cyber}
\resizebox{\textwidth}{!}{%
\begin{tabular}{@{}l ccc ccc cccccc c@{}}
\toprule
 & \multicolumn{3}{c}{\textbf{WMDP}} & \multicolumn{3}{c}{\textbf{CyberSecEval3}} & \multicolumn{6}{c}{\textbf{SOSBench}} & \\
\cmidrule(lr){2-4} \cmidrule(lr){5-7} \cmidrule(lr){8-13} \cmidrule(lr){14-14}
\textbf{Model} & \textbf{bio} & \textbf{chem} & \textbf{cyber} & \textbf{mitre} & \textbf{p.inj} & \textbf{interp.} & \textbf{bio} & \textbf{chem} & \textbf{med} & \textbf{pharm} & \textbf{phys} & \textbf{psy} & \textbf{Avg} \\
\midrule
\multicolumn{14}{@{}l}{\textit{External guardrails}} \\
Qwen3Guard-Gen-8B & 0.085 & 0.463 & 0.156 & 0.876 & 0.615 & 0.962 & \textbf{0.951} & \textbf{0.959} & \textbf{0.935} & 0.951 & 0.900 & \textbf{1.000} & 0.738 \\
Qwen3Guard-Stream-8B & 0.036 & 0.241 & 0.070 & 0.773 & 0.421 & 0.956 & 0.884 & 0.859 & 0.777 & 0.807 & 0.755 & 0.989 & 0.631 \\
WildGuard-7B & 0.111 & 0.341 & 0.207 & 0.860 & 0.475 & 0.956 & 0.929 & 0.839 & 0.908 & 0.792 & 0.862 & 0.983 & 0.689 \\
PolyGuard-Qwen-7B & 0.124 & 0.623 & 0.240 & 0.888 & 0.615 & 0.967 & 0.951 & 0.953 & 0.897 & 0.807 & 0.958 & 0.989 & 0.751 \\
YuFeng-XGuard-8B & 0.210 & 0.730 & 0.232 & 0.844 & 0.615 & 0.962 & 0.935 & 0.902 & 0.880 & 0.921 & 0.835 & 0.989 & 0.755 \\
\midrule
\multicolumn{14}{@{}l}{\textit{Native guardrails}} \\
GradSafe (Llama-2-7B) & 0.015 & 0.027 & 0.056 & 0.575 & 0.333 & 0.862 & 0.767 & 0.344 & 0.619 & 0.222 & 0.644 & 0.818 & 0.440 \\
SIREN (Llama-3.1-8B) & \textbf{0.626} & \textbf{0.793} & 0.301 & 0.751 & 0.364 & 0.889 & 0.872 & 0.745 & 0.856 & 0.927 & 0.814 & 0.936 & 0.740 \\
NExT-Guard (Qwen3-8B) & 0.156 & 0.302 & 0.134 & 0.811 & 0.125 & 0.950 & 0.775 & 0.688 & 0.646 & 0.413 & 0.555 & 0.803 & 0.530 \\
\midrule
\system-lasttok (Qwen3-8B) & 0.071 & 0.341 & 0.312 & \textbf{0.950} & 0.448 & 0.956 & 0.842 & 0.948 & 0.925 & 0.871 & 0.924 & 0.904 & 0.708 \\
\system-gru (Qwen3-8B) & 0.262 & 0.569 & 0.316 & 0.908 & 0.421 & 0.962 & 0.848 & 0.812 & 0.756 & 0.785 & 0.875 & 0.936 & 0.704 \\
\system-ema (Qwen3-8B) & 0.211 & 0.479 & 0.331 & 0.922 & 0.548 & \textbf{0.979} & 0.896 & 0.902 & 0.897 & 0.909 & 0.894 & 0.995 & 0.747 \\ 
\system-lasttok (Qwen3.5-9B) & 0.546 & 0.752 & \textbf{0.583} & {0.942} & \textbf{0.696} & 0.967 & 0.907 & 0.890 & 0.874 & {0.979} & \textbf{0.969} & 0.890 & \textbf{0.833} \\
\system-gru (Qwen3.5-9B) & 0.497 & 0.741 & 0.459 & 0.851 & 0.571 & 0.962 & 0.829 & 0.920 & 0.818 & \textbf{0.984} & 0.906 & 0.929 & 0.789 \\
\system-ema (Qwen3.5-9B) & 0.245 & 0.610 & 0.546 & 0.952 & 0.657 & 0.979 & 0.878 & 0.819 & 0.831 & 0.968 & 0.842 & 0.954 & 0.773 \\ 
\bottomrule
\end{tabular}%
}
\end{table}


\paragraph{Multilingual Safety Evaluation.}
Table~\ref{tab:multilingual} shows competitive detection across the evaluated languages, with the strongest method varying by benchmark and stage. On PolyGuardPrompts, PolyGuard-Qwen-7B leads query classification at 0.869, and YuFeng-XGuard-8B leads response classification at 0.793. Among native guardrails, \system-ema leads both stages, reaching 0.835 for queries on Qwen3.5-9B and 0.766 for responses on Qwen3-8B. On RTP-LX queries, \system-ema (Qwen3.5-9B) is the strongest native entry at 0.847, below Qwen3Guard-Gen-8B (0.877). On RTP-LX responses, \system-lasttok (Qwen3-8B) achieves the highest $F_1$ of 0.898. For multimodal queries on Lingua-SafetyBench, \system-gru (Qwen3.5-9B) leads at 0.788, compared with 0.761 for HiddenDetect and 0.509 for Llama~Guard~4.

\begin{table}[t]
\centering
\caption{Multilingual safety evaluation for queries (Q) and responses (R). PolyGuardPrompts and RTP-LX contain text; Lingua-SafetyBench contains multimodal inputs. Stage indicates the supported decision stages: Q+R, query and response; {R, response only;} p+d, prefill and decode; pre, prefill only; dec, decode only.}
\label{tab:multilingual}
\small
\begin{tabular}{@{}llccccc@{}}
\toprule
 & & \multicolumn{2}{c}{PolyGuardPrompts} & \multicolumn{2}{c}{RTP-LX} & Lingua \\
\cmidrule(lr){3-4} \cmidrule(lr){5-6} \cmidrule(lr){7-7}
Model & Stage & Q & R & Q & R & Q \\
\midrule
\multicolumn{7}{@{}l}{\textit{External guardrails}} \\
Qwen3Guard-Gen-8B & Q+R & 0.865 & 0.781 & \textbf{0.877} & 0.793 & -- \\
Qwen3Guard-Stream-8B & Q+R & 0.865 & 0.770 & 0.852 & 0.855 & -- \\
WildGuard-7B & Q+R & 0.726 & 0.649 & 0.589 & 0.821 & -- \\
PolyGuard-Qwen-7B & Q+R & \textbf{0.869} & 0.749 & 0.833 & 0.769 & -- \\
YuFeng-XGuard-8B & Q+R & 0.857 & \textbf{0.793} & 0.859 & 0.889 & -- \\
{SCM-7B (Qwen2.5-7B)} & {R} & {--} & {0.675} & {--} & {0.718} & {--} \\
Llama~Guard~4 (12B) & Q+R & -- & -- & -- & -- & 0.509 \\
LlavaGuard-7B & Q & -- & -- & -- & -- & 0.273 \\
\midrule
\multicolumn{7}{@{}l}{\textit{Native guardrails}} \\
GradSafe (Llama2-7B) & pre & 0.305 & -- & 0.351 & -- & -- \\
SIREN (Llama-3.1-8B) & p+d & 0.360 & 0.346 & 0.452 & 0.502 & -- \\
NExT-Guard (Qwen3-8B) & p+d & 0.753 & 0.468 & 0.359 & 0.723 & -- \\
HiddenDetect (Llava-1.6-7B) & pre & -- & -- & -- & -- & 0.761 \\
\midrule
\system-lasttok (Qwen3-8B) & p+d & 0.816 & 0.685 & 0.805 & \textbf{0.898} & -- \\
\system-gru (Qwen3-8B) & p+d & 0.830 & 0.748 & 0.806 & 0.851 & -- \\
\system-ema (Qwen3-8B) & p+d & 0.835 & 0.766 & 0.835 & 0.834 & -- \\ 
\system-lasttok (Qwen3.5-9B) & p+d & 0.825 & 0.685 & 0.812 & 0.860 & 0.627 \\
\system-gru (Qwen3.5-9B) & p+d & 0.834 & 0.743 & 0.835 & 0.888 & 0.788 \\
\system-ema (Qwen3.5-9B) & p+d & 0.834 & 0.742 & 0.847 & 0.844 & 0.753 \\  
\bottomrule
\end{tabular}
\end{table}

\subsection{Low Over-Refusal on Benign Traffic} \label{subsec:exp_benign}

\paragraph{Over-Refusal.}
Table~\ref{tab:overrefusal} reports over-refusal rates on XSTest and OR-Bench, measuring the cost of false blocks on benign content. \system-ema (Qwen3-8B) has an average query-side rate of 6.12\%, within the range of the external baselines. Other \system variants reach 13.63--15.35\%, below SIREN (22.41\%) and NExT-Guard (17.38\%). For responses, \system-ema (Qwen3.5-9B) achieves the lowest native average of 4.20\%. Both sequence variants reduce response-side over-refusal relative to the stateless variant on each backbone.


\begin{table}[t]
\centering
\caption{Evaluation of over-refusal rates on XSTest, OR-Bench (\%). Best performance in \textbf{bold}. R denotes response-only classification.}
\label{tab:overrefusal}
\small
\begin{tabular}{@{}llccccccc@{}}
\toprule
 & & \multicolumn{3}{c}{Query} & \multicolumn{3}{c}{Response} \\
\cmidrule(lr){3-5} \cmidrule(lr){6-8}
Model & Stage & XSTest & OR-Bench & Avg & XSTest & OR-Bench & Avg \\
\midrule
\multicolumn{8}{@{}l}{\textit{External guardrails}} \\
Qwen3Guard-Gen-8B & Q+R & 12.88 & 7.71 & 10.29 & 1.81 & 1.44 & 1.63 \\
Qwen3Guard-Stream-8B & Q+R & 10.46 & 4.50 & 7.48 & 0.76 & \textbf{0.13} & 0.45 \\
WildGuard-7B & Q+R & \textbf{1.07} & 8.64 & 4.86 & \textbf{0.38} & 0.24 & \textbf{0.31} \\
PolyGuard-Qwen-7B & Q+R & 3.82 & 8.72 & 6.27 & 21.75 & 1.01 & 11.38 \\
YuFeng-XGuard-8B & Q+R & 7.65 & \textbf{0.30} & \textbf{3.97} & 0.71 & 1.25 & 0.98 \\
SCM-7B (Qwen2.5-7B) & R & -- & -- & -- & 6.33 & 6.14 & 6.24 \\
\midrule
\multicolumn{8}{@{}l}{\textit{Native guardrails}} \\
GradSafe (Llama2-7B) & pre & 14.89 & 5.93 & 10.41 & -- & -- & -- \\
SIREN (Llama-3.1-8B) & p+d & 25.75 & 19.07 & 22.41 & 7.66 & 3.11 & 5.39 \\
NExT-Guard (Qwen3-8B) & p+d & 18.51 & 16.24 & 17.38 & 58.40 & 29.87 & 44.13 \\
\midrule
\system-lasttok (Qwen3-8B) & p+d & 14.55 & 16.15 & 15.35 & 1.86 & 12.33 & 7.10 \\
\system-gru (Qwen3-8B) & p+d & 17.77 & 11.83 & 14.80 & 1.71 & 7.28 & 4.50 \\
\system-ema (Qwen3-8B) & p+d & 10.06 & 2.18 & 6.12 & 2.43 & 6.52 & 4.48 \\ 
\system-lasttok (Qwen3.5-9B) & p+d & 14.49 & 17.69 & 16.09 & 2.19 & 10.57 & 6.38 \\
\system-gru (Qwen3.5-9B) & p+d & 15.96 & 11.46 & 13.71 & 1.67 & 7.18 & 4.43 \\
\system-ema (Qwen3.5-9B) & p+d & 14.49 & 12.77 & 13.63 & 1.38 & 7.02 & 4.20 \\ 
\bottomrule
\end{tabular}
\end{table}

\paragraph{Real-World Benign Traffic.}
Tables~\ref{tab:traffic-text} and~\ref{tab:traffic-vlm} report interception rates on benign text and multimodal traffic.
On text, \system variants intercept an average of 3.69--5.05\% of queries and 2.32--2.80\% of responses. Query-side rates are lower than those of both Qwen3Guard variants, whereas response-side rates are broadly comparable to external baselines. GRU and EMA show comparable interception rates across backbones. On multimodal traffic, average interception rates range from 0.11\% to 0.60\% for queries and from 0.07\% to 0.18\% for responses.

\begin{table}[t]
\centering
\caption{Interception rates on real-world benign text benchmarks (\%),  reported separately for the query (Q) and response (R) sides. Best performance noted in \textbf{bold}. R denotes response-only classification.}
\label{tab:traffic-text}
\small
\begin{tabular}{@{}llcccccc@{}}
\toprule
 & & \multicolumn{3}{c}{Query (Q)} & \multicolumn{3}{c}{Response (R)} \\
\cmidrule(lr){3-5} \cmidrule(lr){6-8}
Model & Stage & Alpaca & WildChat & Avg & Alpaca & WildChat & Avg \\
\midrule
\multicolumn{8}{@{}l}{\textit{External guardrails}} \\
Qwen3Guard-Gen-8B & Q+R & 0.43 & 13.02 & 6.73 & 0.46 & 3.94 & 2.20 \\
Qwen3Guard-Stream-8B & Q+R & 0.39 & 9.95 & 5.17 & \textbf{0.12} & 1.94 & 1.03 \\
WildGuard-7B & Q+R & 0.62 & 5.47 & 3.05 & 0.35 & \textbf{1.43} & \textbf{0.89} \\
PolyGuard-Qwen-7B & Q+R & 0.55 & 14.98 & 7.77 & 1.15 & 10.42 & 5.79 \\
YuFeng-XGuard-8B & Q+R & 0.24 & 10.32 & 5.28 & 0.15 & 4.16 & 2.16 \\
SCM-7B (Qwen2.5-7B) & R & -- & -- & -- & 1.48 & 9.06 & 5.27 \\
\midrule
\multicolumn{8}{@{}l}{\textit{Native guardrails}} \\
GradSafe (Llama2-7B) & pre & \textbf{0.14} & \textbf{1.45} & \textbf{0.79} & -- & -- & -- \\
SIREN (Llama-3.1-8B) & p+d & 3.59 & 6.26 & 4.93 & 1.67 & 3.54 & 2.61 \\
NExT-Guard (Qwen3-8B) & p+d & 0.16 & 5.42 & 2.79 & 0.43 & 8.10 & 4.27 \\
\midrule
\system-lasttok (Qwen3-8B) & p+d & 0.30 & 8.63 & 4.46 & 0.17 & 5.44 & 2.80 \\
\system-gru (Qwen3-8B) & p+d & 0.31 & 7.07 & 3.69 & 0.15 & 4.62 & 2.39 \\
\system-ema (Qwen3-8B) & p+d & 0.24 & 7.43 & 3.83 & 0.21 & 4.43 & 2.32 \\ 
\system-lasttok (Qwen3.5-9B) & p+d & 0.56 & 9.55 & 5.05 & 0.15 & 5.44 & 2.79 \\
\system-gru (Qwen3.5-9B) & p+d & 0.54 & 7.04 & 3.79 & 0.24 & 4.69 & 2.47 \\
\system-ema (Qwen3.5-9B) & p+d & 0.37 & 7.14 & 3.76 & 0.14 & 5.16 & 2.65 \\ 
\bottomrule
\end{tabular}
\end{table}

\begin{table}[t]
\centering
\caption{Interception rates on real-world benign multimodal benchmarks (\%), for the query (Q) and response (R) sides. Best performance noted in \textbf{bold}. }
\label{tab:traffic-vlm}
\resizebox{\textwidth}{!}{%
\begin{tabular}{@{}llcccccccc@{}}
\toprule
 & & \multicolumn{4}{c}{Query (Q)} & \multicolumn{4}{c}{Response (R)} \\
\cmidrule(lr){3-6} \cmidrule(lr){7-10}
Model & Stage & ShareGPT4V & LLaVA-Instr. & MMStar & Avg & ShareGPT4V & LLaVA-Instr. & MMStar & Avg \\
\midrule
\multicolumn{10}{@{}l}{\textit{External guardrails}} \\
Llama Guard 4 (12B) & Q+R & 0.74 & 0.01 & 0.67 & 0.47 & 0.67 & 0.04 & 1.33 & 0.68 \\
LlavaGuard-7B & Q & 1.19 & 1.11 & 0.67 & 0.99 & -- & -- & -- & -- \\
\midrule
\multicolumn{10}{@{}l}{\textit{Native guardrails}} \\
HiddenDetect (LV-1.6-7B) & pre & 1.20 & 0.21 & 0.00 & 0.47 & -- & -- & -- & -- \\
\system-lasttok (Qwen3.5-9B) & p+d & 0.60 & 0.21 & 1.00 & 0.60 & 0.19 & 0.04 & 0.00 & \textbf{0.07} \\
\system-gru (Qwen3.5-9B) & p+d & 0.38 & 0.08 & 0.00 & 0.15 & 0.18 & 0.04 & 0.00 & \textbf{0.07} \\
\system-ema (Qwen3.5-9B) & p+d & 0.26 & 0.06 & 0.00 & \textbf{0.11} & 0.19 & 0.03 & 0.33 & 0.18 \\ 
\bottomrule
\end{tabular}%
}
\end{table}

\subsection{Early-Interception Analysis} \label{subsec:exp_streaming}

The preceding evaluations measure sample-level classification, allowing comparison with detectors that assess completed responses. We next examine when \system first detects an unsafe response during generation.

For each detected unsafe response in the response-side benchmarks, we replay the sequence variant's token-level risk trajectory. The interception point is the first position where the score crosses the threshold, expressed as a fraction of the full response length. We aggregate these fractions across benchmarks. A detector applied only after completion has an interception point of $1.0$.



\begin{figure}[t]
  \centering
  \begin{subfigure}[t]{0.55\linewidth}
    \centering
    \includegraphics[width=\linewidth]{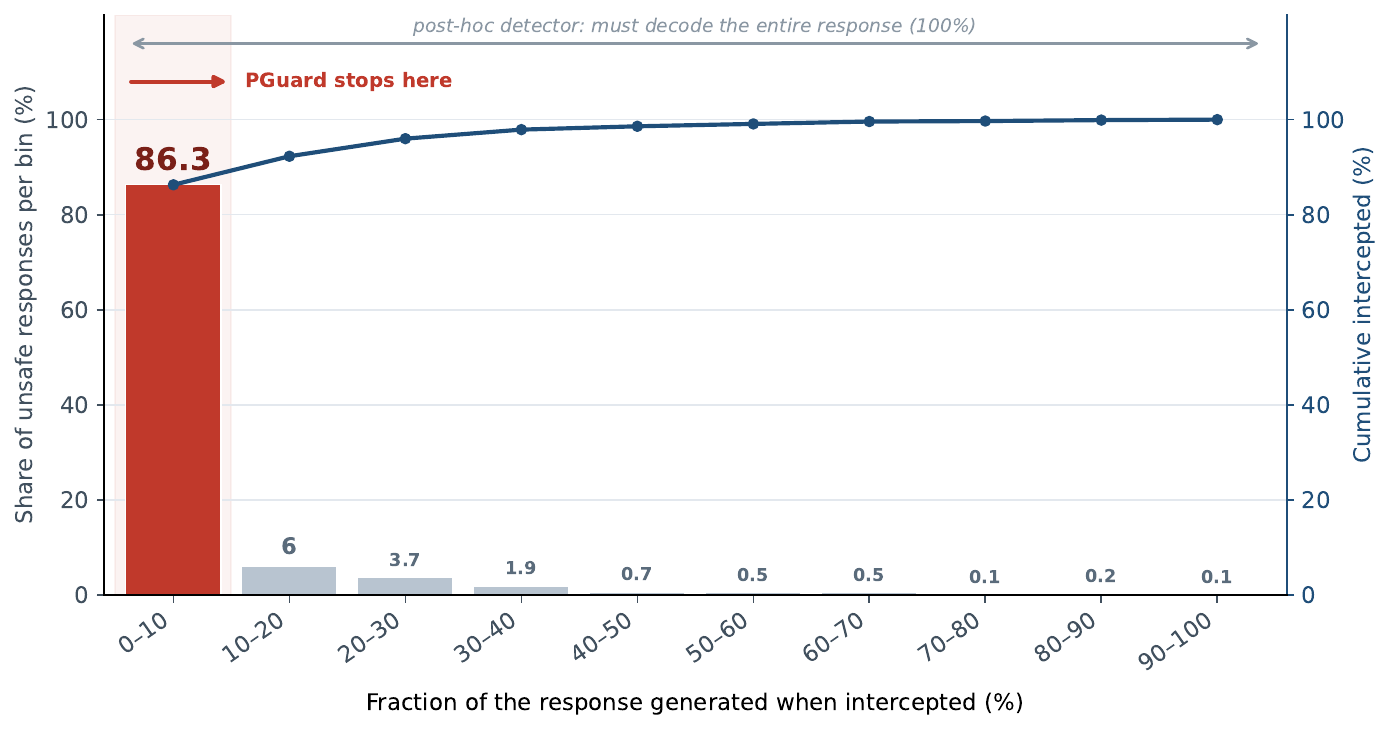}
    \caption{\textbf{First detection of unsafe responses by \system.} Bars show the fraction of detected unsafe responses first flagged in each decile of response length. The line shows the cumulative fraction.}
    \label{fig:early-interception}
  \end{subfigure}
  \hfill
  \begin{subfigure}[t]{0.40\linewidth}
    \centering
    \includegraphics[width=\linewidth]{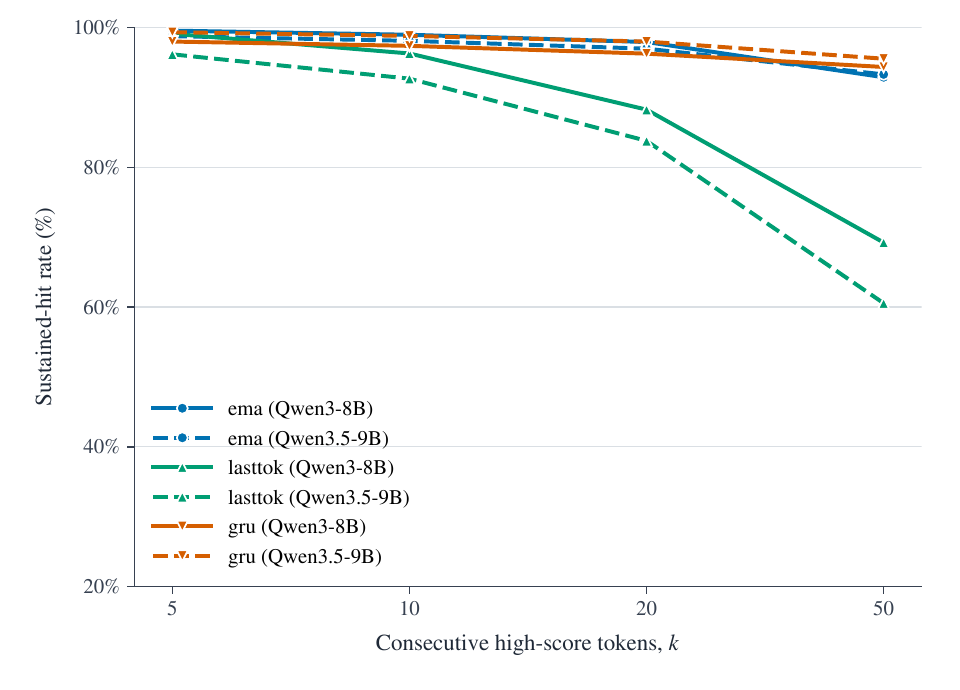}
    \caption{\textbf{Prediction stability of different guardrail variants.} For each $k \in \{5,10,20,50\}$, we report the proportion of flagged samples containing at least $k$ consecutive tokens with risk scores above $0.5$.}
    \label{fig:stability}
  \end{subfigure}
  \caption{Early interception and prediction stability of \system.}
  \label{fig:early-and-stability}
\end{figure}

Figure~\ref{fig:early-interception} shows that 86.3\% of detected unsafe responses are flagged within the first 10\% of generated tokens, and 92.3\% within the first 20\%. These early flags allow generation to stop well before completion. With $p=10$, the wait from any token to the next scoring point is at most nine decode steps. This brief scheduling delay is acceptable for monitoring risk that develops over phrases and passages, while avoiding unnecessary scoring at every token.

\paragraph{Prefix stability.} 
We evaluate prediction stability on S-Eval by measuring, among samples flagged by the guardrail, the proportion containing at least $k$ consecutive tokens with risk scores above $0.5$, indicating sustained predictions of unsafe content. We consider $k \in \{5,10,20,50\}$. As shown in Figure~\ref{fig:stability}, the ema-based variants maintain high stability as the required consecutive span increases: over 95\% of threshold-hit samples sustain above-threshold scores for at least 20 tokens, and approximately 93\% do so for 50 tokens. In contrast, the stateless variants show a more pronounced decline, with only 60--69\% of samples sustaining the threshold for 50 tokens. The results suggest that temporal modeling produces more persistent risk predictions than relying solely on the last-token representation, particularly over longer spans.


\subsection{Signal Analysis} \label{subsec:exp_analysis}

\begin{figure}[t]
  \centering
  \includegraphics[width=\linewidth]{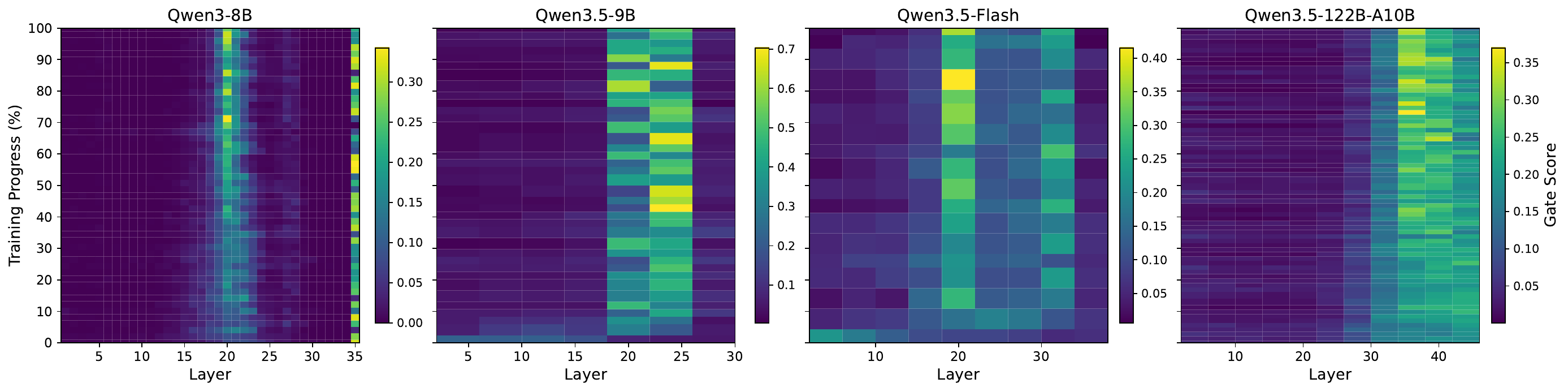}
\caption{\textbf{Learned layer-gate profiles across backbones.} Each panel shows gate weights against relative transformer depth (horizontal) and training progress
   (vertical). Brighter bands indicate greater weight assigned to a layer.}
  \label{fig:layer-screening}
\end{figure}

\paragraph{Layer-gate profiles.}
Figure~\ref{fig:layer-screening} shows how learned layer-gate weights evolve during training. The weights concentrate on different  layer subsets across backbones, supporting backbone-specific selection of capture layers.

\paragraph{Modeling variants on the spectrum.} Table~\ref{tab:pguard-summary}  summarizes performance across backbones and representation-modeling variants.  The stateless variant remains competitive on several tasks, including cyber and CBRN risk classification, whereas sequence modeling generally improves response detection. On Qwen3.5-9B, GRU and EMA-attention achieve text-response $F_1$ scores of 0.860 and 0.833, respectively, compared with 0.795 for the stateless variant. For multimodal responses, the corresponding scores are 0.798, 0.772, and 0.644. GRU performs better on these response tasks, but requires a recurrent state update at every decode step. EMA-attention supports parallel prefill computation and remains the default sequence variant (Section~\ref{subsec:impl_coopt}). The choice therefore depends on the task and serving constraints.


\subsection{Generalization Across Backbones} \label{subsec:exp_backbones}

Table~\ref{tab:pguard-summary} covers Qwen backbones from 8B to 122B total parameters, including dense and MoE architectures and  text-only and multimodal models. The same training recipe applies across these backbones, with layer selection and guard training performed separately for each backbone. These results highlight the applicability of the training recipe across model scales, with the benefits of larger backbones varying by task.


\begin{table}[t]
\centering
\caption{Safety performance (macro-average $F_1$) across backbones and representation-modeling variants.  The highest value in each column within each variant group is shown in \textbf{bold}.}
\label{tab:pguard-summary}
\resizebox{\textwidth}{!}{%
\begin{tabular}{@{}lccccccccc@{}}
\toprule
\multirow{2}{*}{Backbone} & \multicolumn{4}{c}{General safety}
  & \multirow{2}{*}{Jailbreak}
  & \multirow{2}{*}{Cyber\,\&\,CBRN}
  & \multicolumn{3}{c}{Multilingual} \\
\cmidrule(lr){2-5}\cmidrule(lr){8-10}
 & Text Q & Text R & Img Q & Img R
   & & & Text Q & Text R & Img Q \\
\midrule
\multicolumn{10}{@{}l}{\textit{lasttok}} \\
Qwen3-8B & 0.824 & 0.815 & -- & -- & 0.906 & 0.708 & 0.811 & 0.792 & -- \\
Qwen3.5-9B    & 0.822 & 0.795 & 0.635 & 0.644 & 0.917 & \textbf{0.833} & 0.819 & 0.773 & 0.627 \\
Qwen3.5-35B-A3B & 0.822 & 0.812 & 0.734 & 0.658 & 0.941 & 0.816 & 0.754 & 0.791 & 0.756 \\
Qwen3.5-122B-A10B & \textbf{0.838} & \textbf{0.819} & \textbf{0.798} & \textbf{0.730} & \textbf{0.950} & 0.765 & \textbf{0.839} & \textbf{0.803} & \textbf{0.863} \\
\midrule
\multicolumn{10}{@{}l}{\textit{gru}} \\
Qwen3-8B & 0.830 & 0.845 & -- & -- & 0.930 & 0.704 & 0.818 & 0.790 & -- \\
Qwen3.5-9B    & 0.836 & 0.860 & 0.735 & 0.798 & 0.937 & \textbf{0.789} & 0.835 & \textbf{0.816} & \textbf{0.788} \\
Qwen3.5-35B-A3B & 0.846 & \textbf{0.863} & \textbf{0.770} & {\textbf{0.811}} & 0.939 & 0.752 & \textbf{0.857} & 0.806 & 0.782 \\
Qwen3.5-122B-A10B & \textbf{0.851} & 0.857 & 0.745 & 0.805 & \textbf{0.944} & 0.761 & 0.848 & 0.814 & 0.774 \\
\midrule
\multicolumn{10}{@{}l}{\textit{ema}} \\
Qwen3-8B & 0.827 & \textbf{0.858} & -- & -- & 0.946 & 0.747 & 0.835 & 0.800 & -- \\  
Qwen3.5-9B    & \textbf{0.836} & 0.833 & 0.741 & 0.772 & 0.939 & 0.773 & 0.841 & 0.793 & 0.753 \\ 
Qwen3.5-35B-A3B & 0.832 & 0.853 & 0.796 & \textbf{0.795} & \textbf{0.957} & 0.807 & 0.846 & 0.814 & \textbf{0.875} \\ 
Qwen3.5-122B-A10B  & 0.835 & 0.839 & \textbf{0.802} & 0.752 & 0.939 & \textbf{0.812} & \textbf{0.847} & \textbf{0.818} & 0.848 \\
\bottomrule
\end{tabular}%
}
\end{table}

\subsection{Serving Efficiency} \label{subsec:exp_efficiency}

\paragraph{Setup.}
{ We measure the serving cost of \system with SGLang on L20Z GPUs. Appendix~\ref{app:serving_config} lists the serving configuration of each backbone in Table~\ref{tab:eff_setup}. Each run uses a concurrency of 8 and 256 requests per scenario. The input length is 2048, 4096, or 8192 tokens, and the output length is 512 tokens. Each configuration is paired with a run of the bare backbone on the same node, and the two runs alternate in order across 4 rounds. We report the relative increase of time to first token (TTFT) and time per output token (TPOT), together with the relative loss of output throughput. The bare backbone and \system use the same memory fraction for static allocation.\par}

\paragraph{End-to-end overhead.}
{ Table~\ref{tab:eff_e2e} reports TTFT, TPOT, and output throughput changes with each backbone's best combination of optimizations across three backbones and input lengths of 2048--8192 tokens. Among the available measurements, the largest increases are 3.56\% for TTFT and 3.07\% for TPOT. For the stateless variant, the maxima are 1.54\% and 1.78\%, respectively; for EMA-attention, they are 3.56\% and 2.40\%. The GRU variant has maxima of 1.38\% and 3.07\%. Output throughput drops by at most 1.73\% for the stateless variant and 2.26\% for EMA-attention. Most entries stay below 2.5\%, and all entries stay below 3.6\%. This overhead is small and acceptable for large-scale production serving.\par}

\begin{table}[t]
\centering
\caption{Relative changes in TTFT, TPOT, and output throughput (Thpt.) of \system compared with the bare backbone, with an output length of 512 tokens. Each backbone uses its best combination of optimizations. For TTFT and TPOT, positive values indicate increased latency. For throughput, positive values indicate reduced output throughput. Lower is better for all metrics.}
\label{tab:eff_e2e}
\footnotesize
\setlength{\tabcolsep}{4pt}
\begin{tabular}{@{}llccccccccc@{}}
\toprule
\multirow{2}{*}{Backbone} & \multirow{2}{*}{Input}
  & \multicolumn{3}{c}{Stateless}
  & \multicolumn{3}{c}{EMA-attention}
  & \multicolumn{3}{c}{{GRU}} \\
\cmidrule(lr){3-5}\cmidrule(lr){6-8}\cmidrule(lr){9-11}
 & & \makecell{TTFT\\(\%)} & \makecell{TPOT\\(\%)} & \makecell{Thpt.\\(\%)}
   & \makecell{TTFT\\(\%)} & \makecell{TPOT\\(\%)} & \makecell{Thpt.\\(\%)}
   & \makecell{TTFT\\(\%)} & \makecell{TPOT\\(\%)} & \makecell{Thpt.\\(\%)} \\
\midrule
\multirow{3}{*}{\makecell[l]{Qwen3.5-9B\\(TP=1)}}
  & 2048 & +1.43 & +1.78 & +1.73 & +3.56 & +1.69 & +1.76 & +0.73 & +1.64 & +1.55\\
  & 4096 & +1.54 & +1.47 & +1.45 & +3.08 & +1.72 & +1.81 & +0.88 & +1.59 & +1.48\\
  & 8192 & +0.04 & +1.29 & +1.07 & +2.48 & +1.80 & +1.87 & +0.69 & +1.43 & +1.28\\
\midrule
\multirow{3}{*}{\makecell[l]{Qwen3.5-35B-A3B\\(TP=2)}}
  & 2048 & -0.34 & +0.95 & +0.82 & +2.08 & +1.19 & +1.19 & +1.38 & +3.07 & +2.89\\
  & 4096 & +0.25 & -0.93 & -0.88 & +2.15 & +1.42 & +1.43 & +0.97 & +2.18 & +2.02\\
  & 8192 & +0.17 & +0.17 & +0.11 & +1.91 & +1.91 & +1.86 & +0.77 & +2.14 & +1.91\\
\midrule
\multirow{3}{*}{\makecell[l]{Qwen3.5-122B-A10B\\(TP=4)}}
  & 2048 & +0.22 & +1.23 & +0.82 & +0.51 & +2.37 & +2.21 & +0.73 & +1.32 & +1.41\\
  & 4096 & +0.07 & +1.30 & +1.18 & +1.53 & +2.40 & +2.26 & +0.61 & +1.78 & +1.65\\
  & 8192 & +0.29 & +0.99 & +0.89 & +1.20 & +2.31 & +2.11 & +0.63 & +1.79 & +1.51\\
\bottomrule
\end{tabular}
\end{table}

\paragraph{Ablation.}
{ Tables~\ref{tab:eff_ablation_stateless} and~\ref{tab:eff_ablation_ema} add the optimizations one at a time in the listed order. The full ladders reduce output throughput loss from 2.42--3.67\% to 0.16--1.21\% for the stateless variant and from 4.55--5.28\% to 1.53--1.99\% for EMA-attention. Selective decode scoring improves throughput across all three backbones, lowering throughput loss by 0.41--2.43 percentage points for the stateless variant and 0.92--2.19 percentage points for EMA-attention. The gains of the fusion and asynchronous copy steps depend on the backbone.\par}

\begin{table}[t]
\centering
\caption{Cumulative ablation of the stateless variant. Each cell reports the output throughput loss (\%) relative to the bare backbone.}
\label{tab:eff_ablation_stateless}
\small
\resizebox{0.8\textwidth}{!}{%
\begin{tabular}{@{}lccc@{}}
\toprule
Configuration & Qwen3.5-9B & Qwen3.5-35B-A3B & Qwen3.5-122B-A10B \\
\midrule
Unoptimized                & 2.42 & 3.67 & 2.42 \\
\quad + Merge fusion       & 2.39 & 3.69 & 2.49 \\
\quad + Adapter fusion     & 2.48 & 3.54 & 2.29 \\
\quad + Classifier fusion  & 1.50 & 2.10 & 1.17 \\
\quad + Async risk copy    & 1.62 & 2.59 & 1.50 \\
\quad + Selective decode scoring & 1.21 & 0.16 & 0.76 \\
\bottomrule
\end{tabular}}
\end{table}

\begin{table}[t]
\centering
\caption{Cumulative ablation of the EMA-attention variant. Each cell reports the output throughput loss (\%) relative to the bare backbone.}
\label{tab:eff_ablation_ema}
\small
\resizebox{0.8\textwidth}{!}{%
\begin{tabular}{@{}lccc@{}}
\toprule
Configuration & Qwen3.5-9B & Qwen3.5-35B-A3B & Qwen3.5-122B-A10B \\
\midrule
Unoptimized                   & 4.62 & 5.28 & 4.55 \\
\quad + Scan chunk tuning     & 4.58 & 5.73 & 3.93 \\
\quad + Rank-sharded scoring  & --   & 4.66 & 3.65 \\
\quad + QKV fusion            & 4.79 & 4.33 & 4.25 \\
\quad + Classifier fusion     & 4.07 & 0.79 & 3.70 \\
\quad + Decode TP fusion      & --   & 1.68 & 2.52 \\
\quad + Async risk copy       & 4.18 & 3.63 & 2.45 \\
\quad + Selective decode scoring & 1.99 & 1.98 & 1.53 \\
\bottomrule
\end{tabular}}
\end{table}

\paragraph{GRU end-to-end ablation.}
Table~\ref{tab:eff_ablation_gru} reports the end-to-end output throughput loss of the GRU variant relative to the bare backbone, with input/output lengths of 4096/512, concurrency 8, 256 requests, and four paired rounds. Lower values indicate less serving overhead. Async risk copy provides the largest reduction among adjacent reported configurations, bringing throughput loss to 1.16--1.75\% across the three backbones. With selective decode scoring, the loss is further reduced to 0.90--1.09\% across the three backbones.

\begin{table}[t]
\centering
\small
\setlength{\tabcolsep}{3.5pt}
\caption{End-to-end output throughput (TPS) loss (\%) relative to the bare backbone (lower is better), with input/output lengths 4096/512, concurrency 8, 256 requests, and four paired rounds.}
\label{tab:eff_ablation_gru}
\label{tab:gru-end-to-end-ablation}
\resizebox{0.8\textwidth}{!}{%
\begin{tabular}{lrrr}
\toprule
Configuration & Qwen3.5-9B & Qwen3.5-35B-A3B & Qwen3.5-122B-A10B \\
\midrule
GRU baseline (pool prefill) & 3.30 & 4.87 & 3.48 \\
$+$ Vectorized prefill & 3.28 & 4.41 & 3.24 \\
$+$ Layer aggregation fusion & 3.20 & 4.02 & 3.26 \\
$+$ GRU update fusion & 2.91 & 4.17 & 3.21 \\
$+$ Adapter fusion & 2.94 & 4.06 & 3.21 \\
$+$ Async risk copy & 1.16 & 1.75 & 1.29 \\
$+$ Selective decode scoring & 0.90 & 1.09 & 0.92 \\
\bottomrule
\end{tabular}}
\end{table}

\paragraph{Memory overhead.}
{ \system maintains a small memory footprint alongside the backbone. Table~\ref{tab:eff_memory} reports guard weights of {0.03--0.41}\,GiB per GPU, together with per-request state and peak workspace. Peak workspace remains below 0.5\,GiB per GPU for the stateless variant and below 3.1\,GiB per GPU for EMA-attention. The small memory overhead supports large-scale deployment of \system.\par}

\begin{table}[t]
\centering
\caption{Memory overhead of \system.}
\label{tab:eff_memory}
\small
\begin{tabular}{@{}llccc@{}}
\toprule
\multirow{2}{*}{Backbone} & \multirow{2}{*}{Variant}
  & Guard weights & Per-request state & Peak workspace \\
 & & (GiB/GPU) & (MiB/GPU) & (GiB/GPU) \\
\midrule
\multirow{3}{*}{Qwen3.5-9B}
  & Stateless     & 0.06 & --     & 0.19 \\
  & EMA-attention & 0.41 & 0.094  & 2.69 \\
  & GRU           & 0.03 & 0.0016 & 0.69 \\
\midrule
\multirow{3}{*}{Qwen3.5-35B-A3B}
  & Stateless     & 0.04 & --     & 0.09 \\
  & EMA-attention & 0.10 & 0.047  & 1.35 \\
  & GRU           & 0.03 & 0.0016 & 0.34 \\
\midrule
\multirow{3}{*}{Qwen3.5-122B-A10B}
  & Stateless     & 0.04 & --     & 0.14 \\
  & EMA-attention & 0.23 & 0.070  & 2.34 \\
  & GRU           & 0.04 & 0.0017 & 0.60 \\
\bottomrule
\end{tabular}
\end{table}

\subsection{Deployment on the Cost--Accuracy Spectrum} \label{subsec:exp_cascade}

\begin{figure}[t]
\centering
\includegraphics[width=0.8\textwidth]{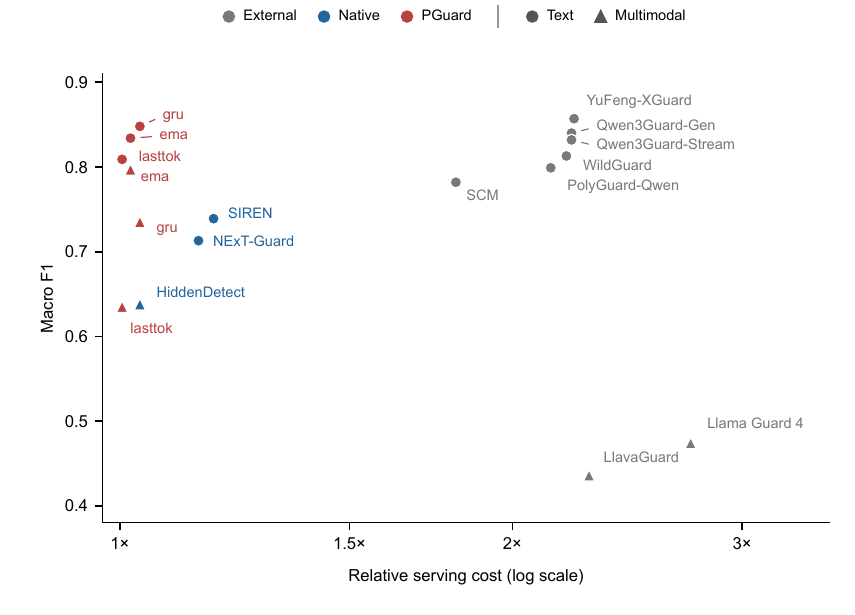}
\caption{The cost--accuracy plane of the evaluated guardrails. Cost is the serving overhead measured with SGLang on the same GPU driven to saturation and normalized by the bare backbone; accuracy is the macro $F_1$ of the corresponding task suites.}
\label{fig:pareto}
\end{figure}

\paragraph{Cost--accuracy overview.} Figure~\ref{fig:pareto}  {summarizes the trade-off between detection accuracy and serving cost. GradSafe is omitted because it requires a backward pass to obtain gradients, which lies outside the serving forward path. Accuracy is measured on} the benchmark suites above{. Cost follows Section~\ref{subsec:impl_overhead}, with each configuration served by SGLang} on the same GPU  {under saturation and normalized to the bare backbone. Serving a second large guard model raises normalized cost to as much as 2.7$\times$. The plotted prior native guardrails achieve 0.64--0.74 macro $F_1$ with 4--18\% overhead. \system combines competitive accuracy with substantially lower serving overhead. On text, \system-gru reaches 0.848 macro $F_1$, exceeding Qwen3Guard-Gen's 0.840, at 1.036$\times$ normalized cost. On multimodal inputs, \system-ema achieves 0.797 macro $F_1$ at 1.019$\times$ cost, compared with 0.638 for the strongest prior native entry. These operating points pair strong detection quality with only 3.6\% and 1.9\% additional serving cost for text and multimodal inputs, respectively.}

\paragraph{Two-tier deployment.} For deployments with a larger accuracy budget, the native variant composes with an external adjudicator instead of replacing it. Table~\ref{tab:guard-compare} evaluates the two-tier deployment  {on S-Eval-Jailbreak, with Qwen3.5-Flash serving as both the \system backbone and the response generator. We regenerate the responses and cascade \system with an external reasoning-based guardrail. Jailbreak prompts are selected to increase the occurrence of risky responses, providing a stress test that approximates risky online traffic.} It reaches the best overall score (96.72 average, versus 95.16 for \system alone), confirming that the native screen and the external adjudicator are complementary. Note that the gain comes entirely from precision, because recall is capped by the first tier. The adjudicator can only review traffic the native screen already flags, and this ceiling is intrinsic to any cascade. The economics of the pattern therefore hinge on the flagged fraction escalated to the second tier.
On the \textsc{Alpaca} traffic set, the native  {guard flags 0.215\% of responses, so only this fraction requires external review. This reduces second-tier calls by approximately 99.8\% compared with reviewing every response. After review, the block rate falls} to 0.056\% (16 samples){. The native guard therefore screens all traffic, while the external adjudicator reviews only flagged responses}.

\begin{table}[t]
\centering
\caption{ {Two-tier guard comparison. Query, Response, and Avg are evaluated on the S-Eval-Jailbreak stress workload, with responses generated by Qwen3.5-Flash.} Each cell reports the
{$F_1$} with its precision/recall breakdown in parentheses, i.e.\
\texttt{score(P/R)}; higher is better. {Avg is computed on the pooled query and response samples. Block Rate is measured separately as the fraction of \textsc{Alpaca} responses blocked.}}
\label{tab:guard-compare}
\resizebox{\textwidth}{!}{%
\begin{tabular}{@{}lcccc@{}}
\toprule
Method & Query & Response & Avg & Block Rate(\%)\\
\midrule
\system {(Qwen3.5-Flash)} & 97.56 (95.24/100.00) & 90.48 (95.00/86.36) & 95.16 (95.16/95.16) & 0.215\\
YuFeng-XGuard         & 95.24 (90.91/100.00) & 88.37 (90.48/86.36) & 92.91 (90.77/95.16) & 0.144\\
\system{}+YuFeng-XGuard & 98.77 (97.56/100.00) & 92.68 (100.00/86.36) & 96.72 (98.33/95.16) & 0.056\\
\bottomrule
\end{tabular}}
\end{table}


\section{Background and Related Work} \label{sec:bg}

\paragraph{External guardrails.}  {External guardrails use a separate model to classify queries and responses. Llama Guard~\citep{llamaguard4,llamaguard3} uses instruction tuning with a content taxonomy, with its fourth generation supporting multimodal input. Related approaches include} ShieldGemma~\citep{shieldgemma}, Granite Guardian~\citep{graniteguardian}, NVIDIA's Nemotron/Aegis safety models~\citep{nemoguard}, and ShieldLM~\citep{shieldlm}. Qwen3Guard~\citep{qwen3guard}  {offers generative and streaming variants. The streaming variant adds} classification heads to a base LLM  {to produce token-level verdicts. Despite its token-level outputs, Qwen3Guard-Stream still requires a separate LLM, adding model memory and inference computation beyond those of the protected backbone. SCM~\citep{scm} likewise runs a separate Qwen2.5-7B model with token-level and sequence-level classification heads to monitor streamed responses, adding its own model memory and inference computation. WildGuard~\citep{wildguardmix} jointly classifies} prompts, responses, and refusal behavior,  {while PolyGuard~\citep{polyguard} extends safety classification to 17 languages. GuardReasoner~\citep{guardreasoner,guardreasonervl} and YuFeng-XGuard~\citep{yufengxguard} generate explicit reasoning before a verdict, adding computation per request. For multimodal moderation, LlavaGuard~\citep{llavaguard} targets images, while SafeWatch~\citep{safewatch} and OmniGuard~\citep{omniguard} extend coverage} to video and omni-modal inputs{. PolicyShiftGuard~\citep{policyshiftguard} addresses} cross-product policy adaptation, and JailGuard~\citep{jailguard}  {detects attacks through} black-box input mutation. These  {approaches require serving an additional model. Response-level detectors also defer decisions until the response is complete}.

\paragraph{Native guardrails: signal sources.}  {Native guardrails instead derive safety signals from the protected model}. GradSafe~\citep{gradsafe} computes gradients  {for} a paired compliant response and scores prompts  {using cosine similarity over} safety-critical parameter slices. HiddenDetect~\citep{hiddendetect}  {identifies unsafe multimodal inputs through refusal-related hidden-state directions before generation; Activation-Cone~\citep{activationcone}} probes the first few generated tokens. SIREN~\citep{siren} trains lightweight per-layer probes over safety neurons,  {whereas kNNGuard~\citep{knnguard} uses a nearest-neighbor bank of last-token activations.} TrajGuard~\citep{trajguard} monitors Mahalanobis distances  {along hidden-state trajectories}, and Oldfield's truncated polynomial classifier~\citep{tpc}  {examines higher-order residual-stream readouts}. NExT-Guard~\citep{nextguard} selects safety-relevant sparse autoencoder features  and scores them token by token{. VLMGuard~\citep{vlmguard} trains an MLP using SVD-derived pseudo-labels, while ARGUS~\citep{argus} combines} per-layer logistic probes with activation-level intervention for multimodal inputs.  {These approaches differ in their signal sources, decision granularity, and serving support, motivating joint evaluation of detection quality and runtime cost}.

\paragraph{Native guardrails: streaming and production.} Streaming  {detection is the setting most closely related} to \system. Segment-Level Coherence~\citep{segcoherence} pools windowed probe evidence for  {token-level decisions.} Stop Early, Spend Less~\citep{stopearly}  {integrates} a probe into the vLLM decode loop for early termination{. Kelp~\citep{kelp} models risk using a continuous-time latent state} trained with response-level labels{, while SafeSwitch~\citep{safeswitch} combines} a probe with a refusal head{. Production systems also combine lightweight screening with more expensive adjudication. Anthropic's Constitutional Classifiers combine} an input classifier with an output value head~\citep{constitutionalclassifiers}{. Its successor uses internal probes trained with sliding-window mean logit smoothing (SWiM) to flag traffic for} a larger exchange classifier~\citep{constclasspp}. Google DeepMind's Gemini probes~\citep{kramar} similarly  {use token-level internal classifiers} and escalate less than 10\% of traffic. High-Stakes Probes~\citep{highstakes}  {studies cascades that route a fixed fraction of traffic from a lightweight probe to an expensive judge. Together, these approaches motivate evaluating native screening as part of a complete serving pipeline}.

\paragraph{Intervention mechanisms.}  {Other methods directly alter generation through} representation reading and control~\citep{repe}, contrastive activation addition~\citep{caa},  {or circuit breakers implemented with LoRA adapters~\citep{circuitbreakers}. Related approaches include learned null-space steering~\citep{alphasteer} and multimodal defenses acting on prompts or activations~\citep{adashield,ecso,autosteer}. \system supplies token-level risk scores and leaves the response policy, such as reporting, interruption, or steering, to the deployment. These intervention mechanisms can therefore complement its detection interface}.

\paragraph{Positioning of \system.}  {\system brings together representation-based safety detection and serving-engine integration, evaluating detection quality alongside runtime overhead. Its stateless and sequence variants offer different accuracy--cost trade-offs within a shared SGLang integration (Section~\ref{sec:impl}). The evaluation spans} dense and MoE backbones,  {text-only} and multimodal models, and parameter scales from 8B to 122B (Section~\ref{subsec:exp_backbones}). {The planned release includes the guard models, evaluation pipeline, and serving integration.}

\section{Conclusion} \label{sec:conclusion}

We presented \system, a native streaming guardrail that operates directly on the intermediate representations of a frozen backbone. Through dynamic layer gating, \system aggregates residual-stream features from selected layers, enabling a trained guard module to produce risk scores throughout both prefill and autoregressive decoding. We further explored three guard configurations: a stateless variant and two sequence-aware variants based on EMA-attention and GRU, offering different trade-offs between detection performance and serving efficiency.

Our evaluation across more than 30 benchmarks demonstrates that \system achieves competitive detection performance against dedicated external guardrails while consistently outperforming existing native baselines in aggregate. Its streaming predictions enable early detection of unsafe responses, allowing potential risks to be identified before generation is complete. Integrated into SGLang, \system incurs only 3.56\% and 3.07\% increases in TTFT and TPOT, respectively, across the evaluated serving configurations. 
These results demonstrate the practical viability of native guardrails for online safety screening, while retaining the option of external adjudication for flagged traffic.
We are also extending the integration to vLLM, with adaptation work currently underway.

\paragraph{Limitations.}
The effectiveness of \system depends on the extent to which safety-relevant information is encoded in the backbone’s intermediate representations. Attacks that induce harmful behavior without producing sufficiently informative signals in the captured representations may evade detection. Deep role-playing and fictional framing are potential examples that warrant targeted investigation. Furthermore, adapting \system to a new backbone requires calibration-set pretraining for layer selection, followed by guard-module training, even when the new backbone belongs to the same model family. These requirements introduce additional costs when deploying \system across different backbones.

Our current jailbreak evaluation relies on static benchmarks and does not establish robustness against adaptive attacks involving multi-turn interactions. Evaluating \system under dynamic red-teaming scenarios is therefore an important direction for future work. In addition, the framework leaves intervention policies to the deployment, motivating further investigation into how streaming risk scores can be translated into effective runtime actions. More expressive sequence models, including recurrent–EMA combinations and full self-attention, also merit exploration when serving budgets permit. Finally, we plan to extend the evaluation to larger backbones and additional modalities, and to release the guard models, benchmark pipeline, and SGLang serving integration alongside this work.

\section{Authors}
\textbf{Core Contributors.}
Junyu Lin*, Mengjie Wu*, Yonghao Dong$\ddagger$, Yuchuan Fu$\dagger$, Jinfeng Li\envelopefootnote{Corresponding author: Jinfeng Li (jinfengli.ljf@alibaba-inc.com)}, Xiaodan Li, Xiaohan Yuan, Yicheng Liu$\ddagger$, Yuefeng Chen, Hui Xue, Jingyi Wang$\dagger$

\textbf{Contributors.}
Yitong Yang, Jiyuan Fu, Qiang He$\ddagger$

{\small{(* equal contribution) \newline ($\dagger$ Intelligent System Security Lab, Zhejiang University, Hangzhou) \newline ($\ddagger$ Services Computing Technology and System Lab, Huazhong University of Science and Technology, Wuhan)}}

\clearpage
\appendix



\section{Multi-Task Taxonomy and Training-Data Composition} \label{app:taxonomy}

\setlength{\tabcolsep}{4pt}
\begin{longtable}{@{}p{0.17\textwidth}p{0.25\textwidth}p{0.5\textwidth}@{}}
\caption{\system safety taxonomy adapted from YuFeng-XGuard~\cite{yufengxguard}:
9 risk dimensions and 29 fine-grained categories, each with its definition.}\\
\label{tab:taxonomy}\\
\toprule
\textbf{Risk dimension} & \textbf{Risk category} & \textbf{Description} \\
\midrule
\endfirsthead
\multicolumn{3}{@{}l}{\tablename~\thetable\ (continued)}\\
\toprule
\textbf{Risk dimension} & \textbf{Risk category} & \textbf{Description} \\
\midrule
\endhead
\midrule
\multicolumn{3}{@{}r}{\textit{continued on next page}}\\
\endfoot
\bottomrule
\endlastfoot

\textbf{A. Crimes \& Illegal Activities}
 & \textit{a1.}Pornographic Contraband & Obscene or sexually explicit material, including sexual depictions and pornographic resources, plus non-consensual or illegal sexual conduct and date-rape drugs. \\
 & \textit{a2.}Drug Crimes & Cultivation, synthesis, trafficking, and glorified personal use of illegal drugs and controlled substances. \\
 & \textit{a3.}Dangerous Weapons & Illicit blades, firearms, and explosives, notably CBRN (chemical, biological, radiological, nuclear) agents and devices such as anthrax, mustard/chlorine, Cs-137/Co-60 dirty bombs, and TATP/RDX car bombs. \\
 & \textit{a4.}Property Infringement & Unlawfully taking or damaging others' property via burglary, theft, fraud, extortion, or robbery. \\
 & \textit{a5.}Economic Crimes & Financial fraud, market manipulation, tax evasion, bribery, money laundering, and IP infringement that disrupt economic order. \\
\addlinespace

\textbf{B. Hate Speech}
 & \textit{b1.}Abusive Curses & Profane, vulgar, or insulting language that degrades or attacks individuals or groups. \\
 & \textit{b2.}Defamation & Circulating false or fabricated claims to damage a person's or organization's reputation. \\
 & \textit{b3.}Threats \& Intimidation & Direct or indirect expressions of intent to harm, coerce, or intimidate others. \\
 & \textit{b4.}Cyberbullying & Sustained online harassment, humiliation, exclusion, or doxxing of an individual. \\
\addlinespace

\textbf{C. Physical \& Mental Health}
 & \textit{c1.}Personal Safety & Content encouraging harm to others or oneself, suicide and self-harm, fighting, domestic violence, or reckless dangerous behavior. \\
 & \textit{c2.}Mental Health & Material that worsens psychological distress or replaces proper care, including emotional manipulation, pseudo-scientific remedies, and discouraging treatment. \\
\addlinespace

\textbf{D. Ethics \& Morality}
 & \textit{d1.}Social Ethics & Breaches of widely held moral norms, including prejudice and discrimination, devaluation of life or animal abuse, and coaching dishonest behavior. \\
 & \textit{d2.}Science \& Tech Ethics & Hazardous use of science and technology, such as AI-enabled lethal weapons, anti-human plans, and unethical biotech like human cloning or germline editing. \\
\addlinespace

\textbf{E. Data Privacy}
 & \textit{e1.}Personal Privacy & Exposing, inferring, or speculating about non-public personal information (identity, contacts, financial, medical, judicial records) or teaching how to harvest it. \\
 & \textit{e2.}Commercial Secret & Disclosing or reconstructing a specific company's non-public internal information (strategy, financials, customers, HR) or instructing how to obtain it. \\
\addlinespace

\textbf{F. Cybersecurity}
 & \textit{f1.}Access Control & Gaining credentials or privileges beyond authorized limits, including stolen auth secrets (AK/SK), device jailbreaking, and privilege escalation. \\
 & \textit{f2.}Malicious Code & Designing, generating, or spreading viruses, worms, Trojan horses, and ransomware. \\
 & \textit{f3.}Hacker Attack & Unauthorized intrusion into, disruption of, or data exfiltration from systems, networks, or services. \\
 & \textit{f4.}Physical Security & Damaging network hardware or critical infrastructure (cables, fiber, base stations, substations) that underlies connectivity. \\
 & \textit{f5.}Social Engineering, Phishing & Deceiving users into revealing information or granting access, via cloned login pages, forged-identity lures, spear-phishing, and filter evasion. \\
\addlinespace

\textbf{G. Extremism}
 & \textit{g1.}Violent Terrorist Activities & Promotion, glorification, or instruction for politically, religiously, or ideologically motivated terrorist attacks and armed conflict. \\
 & \textit{g2.}Social Disruption & Inciting or amplifying hostility, confrontation, or division among groups by ideology, religion, ethnicity, or identity. \\
 & \textit{g3.}Extremist Ideological Trends & Spreading radical ideologies that advocate violence, hatred, or the overthrow of established systems. \\
\addlinespace

\textbf{H. Inappropriate Suggestions}
 & \textit{h1.}Finance & Speculative, unlicensed, or high-stakes financial advice that risks serious monetary loss. \\
 & \textit{h2.}Medicine & Diagnoses, treatments, or medication advice given without professional qualification that could endanger health. \\
 & \textit{h3.}Law & Unauthorized legal interpretation or advice that could lead to adverse legal consequences. \\
\addlinespace

\textbf{I. Risks Involving Minors}
 & \textit{i1.}Corruption of Minors & Leading minors toward harmful or delinquent behavior such as underage smoking, drinking, or truancy. \\
 & \textit{i2.}Minor Abuse \& Exploitation & Depicting, facilitating, or encouraging physical, psychological, or sexual abuse of children. \\
 & \textit{i3.}Minor Delinquency & Involving minors as perpetrators of crime, or coaching them in illegal acts. \\

\end{longtable}

\begin{table}[H]
\centering
\small
\caption{Training-data composition by modality.}
\label{tab:train-data}
\begin{tabular}{@{}llrp{0.40\textwidth}@{}}
\toprule
Modality & Source & \#Samples & Description \\
\midrule
Text & In-house annotated data & 157{,}497 & Safety-annotated data mapped to the shared taxonomy. \\
 & Public safety corpora & 71{,}095 & Train splits of PolyGuardMix~\cite{polyguard}, SafeRLHF~\cite{saferlhf}, Aegis~2.0~\cite{aegis2}, WildGuardMix~\cite{wildguardmix}, BeaverTails~\cite{beavertails}, and SORRY-Bench~\cite{sorrybench}. \\
\cmidrule(l){2-3}
 & \textit{Text total} & 228{,}592 & \\
\midrule
MultiModality
 & SPA-VL (train split) & 78{,}247
   & Safety preference alignment data for vision-language models~\cite{spavl}. \\
 & Honey-Data     & 20{,}000
   & General multimodal SFT corpus~\cite{honeydata} for benign data. \\
\cmidrule(l){2-3}
 & \textit{Image-text total} & 98{,}247 & \\
\bottomrule
\end{tabular}
\end{table}

\par\addvspace{\baselineskip}
\noindent\begin{minipage}{\textwidth}
\section{Decision Threshold Configuration} \label{app:thresholds}


\begin{table}[H]
\centering
\caption{Decision thresholds $\tau_{c,y}$ in percent, calibrated separately
for each guard. Rows are split by stage and
input: Q and R denote the query and response stages, and T and VL denote text
and image-text inputs. Columns are the nine risk dimensions (A--I). Each cell
lists the per-category thresholds of that dimension in \textit{a1}$\to$\textit{an}
order.}
\label{tab:thresholds}
\footnotesize
\setlength{\tabcolsep}{4pt}
\resizebox{\textwidth}{!}{%
\begin{tabular}{@{}llccccccccc@{}}
\toprule
Configuration & Stage ($\tau$, \%) & \textbf{A} Crimes & \textbf{B} Hate & \textbf{C} Health & \textbf{D} Ethics & \textbf{E} Privacy & \textbf{F} Cyber & \textbf{G} Extremism & \textbf{H} Inappr. & \textbf{I} Minors \\
\midrule
\multirow{2}{*}{lasttok (Qwen3-8B)} & Q-T & 65/36/54/42/53 & 66/59/10/18 & 71/47 & 71/69 & 75/74 & 87/62/25/62/10 & 31/48/62 & 68/54/34 & 87/50/24 \\
 & R-T & 55/82/55/19/55 & 33/43/24/20 & 47/65 & 61/14 & 45/70 & 41/65/30/43/29 & 59/44/68 & 77/67/56 & 38/72/12 \\
\multirow{2}{*}{gru (Qwen3-8B)} 
& Q-T  & 75/83/61/28/68 & 73/24/10/10 & 61/47 & 62/56 & 46/74 & 80/33/40/21/19 & 29/70/56 & 70/62/16 & 83/57/18 \\
& R-T  & 72/97/79/21/79 & 43/23/13/15 & 77/99 & 90/35 & 91/89 & 72/45/29/42/68 & 86/70/66 & 87/69/10 & 74/85/19 \\
\multirow{2}{*}{ema (Qwen3-8B)} & Q-T & 45/25/65/19/66 & 42/10/10/10 & 26/35 & 30/11 & 45/35 & 30/39/11/13/12 & 27/66/19 & 49/43/14 & 37/30/10 \\
 & R-T & 45/38/34/37/54 & 34/34/16/12 & 37/41 & 58/19 & 56/69 & 35/59/14/29/10 & 50/31/65 & 44/46/10 & 18/49/10 \\
\multirow{4}{*}{lasttok (Qwen3.5-9B)} 
& Q-T   & 71/25/89/89/48 & 75/11/18/20 & 47/81 & 66/27 & 77/75 & 52/60/30/13/10 & 38/71/64 & 72/45/10 & 92/76/10 \\
& Q-VL  & 90/64/89/95/15 & 56/10/11/15 & 90/11 & 87/10 & 91/53 & 19/10/27/19/10 & 47/29/27 & 10/52/14 & 90/55/10 \\
& R-T  & 74/96/83/54/50 & 54/56/45/51 & 66/70 & 80/44 & 74/87 & 84/70/49/69/40 & 82/62/80 & 89/85/74 & 81/87/11 \\
& R-VL & 83/94/64/84/30 & 39/26/17/25 & 83/72 & 70/35 & 79/39 & 83/54/25/70/10 & 77/90/75 & 76/93/38 & 55/80/10 \\
\multirow{4}{*}{gru (Qwen3.5-9B)}
& Q-T   & 86/41/55/39/62 & 74/30/14/22 & 70/67 & 54/38 & 62/80 & 76/32/63/53/49 & 64/70/33 & 53/64/12 & 82/48/21 \\
& Q-VL  & 11/10/10/10/10 & 10/10/43/11 & 10/10 & 10/14 & 10/10 & 13/19/65/10/10 & 19/11/10 & 10/10/10 & 10/10/10 \\
& R-T   & 65/85/73/32/49 & 26/80/51/70 & 70/54 & 80/38 & 68/91 & 77/37/20/50/42 & 70/65/51 & 90/95/10 & 96/51/10 \\
& R-VL  & 18/10/50/39/10 & 10/10/19/10 & 14/10 & 58/53 & 10/10 & 10/10/10/10/10 & 10/10/10 & 33/95/83 & 10/10/10 \\
\multirow{4}{*}{ema (Qwen3.5-9B)}
& Q-T  & 52/35/74/34/29 & 54/14/10/33 & 40/62 & 54/43 & 63/70 & 25/68/19/10/11 & 20/50/45 & 39/54/11 & 30/63/14 \\
& Q-VL & 10/10/11/10/10 & 10/15/10/10 & 10/10 & 10/22 & 10/10 & 10/10/10/10/10 & 10/39/20 & 10/10/10 & 10/10/10 \\
& R-T  & 73/96/59/72/65 & 36/28/90/54 & 67/94 & 95/25 & 76/88 & 36/81/35/18/81 & 85/65/92 & 88/73/50 & 87/73/20 \\
& R-VL & 30/77/10/61/49 & 30/14/45/27 & 23/20 & 33/63 & 47/10 & 35/25/27/10/86 & 10/20/34 & 38/98/49 & 10/10/10 \\
\multirow{4}{*}{lasttok (Qwen3.5-flash)}
& Q-T   & 75/64/75/61/71 & 73/65/10/10 & 61/62 & 69/42 & 83/85 & 75/73/16/26/10 & 65/25/67 & 64/59/65 & 75/75/18 \\
& Q-VL  & 80/32/54/83/61 & 58/21/10/10 & 83/10 & 85/30 & 75/65 & 19/11/70/40/11 & 65/21/19 & 16/44/16 & 78/72/10 \\
& R-T   & 51/58/53/39/56 & 42/26/21/49 & 61/68 & 61/34 & 59/88 & 57/48/34/51/31 & 60/45/49 & 80/72/38 & 28/69/10 \\
& R-VL  & 93/26/26/65/11 & 17/12/27/11 & 60/19 & 53/10 & 57/24 & 58/10/21/35/10 & 37/28/34 & 52/93/37 & 20/38/10 \\
\multirow{4}{*}{gru (Qwen3.5-flash)}
& Q-T   & 48/76/62/42/42 & 63/25/10/13 & 50/64 & 75/23 & 78/75 & 86/39/45/10/10 & 77/75/58 & 51/72/22 & 77/76/15 \\
& Q-VL  & 10/10/10/10/10 & 12/11/55/10 & 10/10 & 10/10 & 10/10 & 10/27/11/10/10 & 10/10/10 & 10/10/10 & 24/35/10 \\
& R-T   & 72/86/78/63/40 & 21/76/29/40 & 65/30 & 77/53 & 56/83 & 59/38/33/33/54 & 86/49/76 & 87/58/31 & 57/82/10 \\
& R-VL  & 28/50/34/10/10 & 14/15/10/10 & 28/13 & 34/32 & 10/10 & 10/36/10/10/10 & 10/11/10 & 14/94/10 & 10/10/10 \\
\multirow{4}{*}{ema (Qwen3.5-flash)}
& Q-T  & 65/85/67/56/65 & 73/15/11/44 & 75/25 & 75/42 & 62/92 & 77/13/29/10/10 & 45/51/50 & 52/61/10 & 54/56/19 \\
& Q-VL & 10/10/10/10/10 & 24/25/93/14 & 10/10 & 10/15 & 10/10 & 10/44/10/10/10 & 10/46/10 & 12/10/10 & 10/10/10 \\
& R-T  & 65/78/41/28/61 & 47/16/25/39 & 53/94 & 66/19 & 47/81 & 80/59/27/61/55 & 74/45/90 & 86/63/33 & 43/57/18 \\
& R-VL & 40/83/10/64/10 & 17/31/11/15 & 10/32 & 21/87 & 20/16 & 34/10/15/10/95 & 10/19/86 & 11/94/55 & 10/10/10 \\
\multirow{4}{*}{lasttok (Qwen3.5-122B-A10B)}
& Q-T   & 76/84/77/46/61 & 67/31/10/13 & 63/80 & 87/28 & 73/93 & 10/75/72/15/10 & 67/52/63 & 77/45/26 & 34/70/34 \\
& Q-VL  & 10/11/12/20/10 & 35/10/10/10 & 30/10 & 24/10 & 11/10 & 10/10/13/10/10 & 10/10/15 & 10/10/10 & 22/35/10 \\
& R-T   & 64/79/68/18/61 & 36/29/39/14 & 40/85 & 70/17 & 39/85 & 41/60/34/33/13 & 59/43/70 & 72/72/39 & 54/68/10 \\
& R-VL  & 19/36/10/60/10 & 29/19/17/19 & 12/17 & 29/75 & 28/10 & 13/10/19/10/10 & 10/11/11 & 38/87/42 & 10/10/10 \\
\multirow{4}{*}{gru (Qwen3.5-122B-A10B)}
& Q-T   & 71/82/77/78/28 & 79/25/17/22 & 55/56 & 68/33 & 78/83 & 57/40/69/26/15 & 61/59/38 & 84/57/12 & 65/47/10 \\
& Q-VL  & 10/10/10/10/10 & 11/25/10/11 & 10/10 & 10/10 & 10/10 & 10/10/10/10/10 & 10/12/10 & 10/10/10 & 10/21/10 \\
& R-T   & 79/93/56/27/67 & 25/16/28/74 & 61/98 & 74/18 & 52/78 & 42/40/53/38/41 & 71/61/25 & 93/83/66 & 84/30/10 \\
& R-VL  & 10/10/75/16/10 & 12/10/15/10 & 17/10 & 12/54 & 20/10 & 10/31/10/10/10 & 10/15/10 & 32/97/10 & 10/10/10 \\
\multirow{4}{*}{ema (Qwen3.5-122B-A10B)}
& Q-T   & 70/82/46/48/60 & 65/16/10/20 & 59/39 & 75/37 & 69/86 & 52/70/35/65/11 & 52/55/10 & 68/35/10 & 48/59/10 \\
& Q-VL  & 10/10/10/10/10 & 18/33/10/10 & 15/10 & 17/24 & 15/10 & 10/10/13/10/10 & 36/47/16 & 11/10/10 & 11/47/10 \\
& R-T   & 85/81/77/29/59 & 47/10/15/34 & 70/97 & 86/14 & 70/87 & 68/74/30/36/21 & 61/56/80 & 97/74/37 & 49/66/10 \\
& R-VL  & 10/35/10/42/10 & 10/10/12/10 & 10/10 & 47/25 & 77/10 & 20/10/18/10/10 & 12/17/14 & 10/92/10 & 10/10/10 \\
\bottomrule
\end{tabular}%
}
\end{table}

\end{minipage}\par
\addvspace{\baselineskip}
\noindent\begin{minipage}{\textwidth}
\section{Layer Selection Configurations} \label{app:layers}

\begin{table}[H]
\centering
\caption{Top-3 Layer Selection Configurations}
\label{tab:top3_layers}
\begin{tabular}{lll}
\toprule
Modeling Method & Backbone Model & Top-3 Layers \\
\midrule
lasttok & Qwen3-8B & \textbf{17,18,19} \\
gru / ema & Qwen3-8B & \textbf{19,20,35} \\
lasttok / gru / ema & Qwen3.5-9B & \textbf{20, 24, 28} \\
lasttok / gru / ema & Qwen3.5-flash & \textbf{20, 24, 28} \\
ema & Qwen3.5-122B-A10B & \textbf{32,36,40} \\
lasttok / gru & Qwen3.5-122B-A10B & \textbf{36,40,44} \\
\bottomrule
\end{tabular}
\end{table}
\end{minipage}\par


\addvspace{\baselineskip}
\noindent\begin{minipage}{\textwidth}
\section{{Serving Configuration}} \label{app:serving_config}

\begin{table}[H]
\centering
\caption{Serving configuration of each backbone. Memory fractions are listed for the stateless, EMA-attention, and GRU variants, respectively.}
\label{tab:eff_setup}
\small
\begin{tabular}{@{}lcccc@{}}
\toprule
Backbone & TP & SGLang & Capture indices & Memory fraction \\
\midrule
Qwen3.5-9B        & 1 & 0.5.11 & 20, 24, 28 & 0.9 / 0.8 / 0.8 \\
Qwen3.5-35B-A3B   & 2 & 0.5.11 & 20, 24, 28 & 0.9 / 0.8 / 0.8 \\
Qwen3.5-122B-A10B & 4 & 0.5.11 & \makecell[l]{EMA-attention: 32, 36, 40\\Stateless/GRU: 36, 40, 44} & 0.9 / 0.8 / 0.8 \\
\bottomrule
\end{tabular}
\end{table}
\end{minipage}\par

\clearpage
\bibliography{biblio}

\bibliographystyle{colm2024_conference}


\end{CJK} 
\end{document}